\documentclass{article}
\PassOptionsToPackage{numbers, compress}{natbib}
    \usepackage[preprint]{neurips_2023}

\usepackage[utf8]{inputenc} % allow utf-8 input
\usepackage[T1]{fontenc}    % use 8-bit T1 fonts
\usepackage{hyperref}       % hyperlinks
\usepackage{url}            % simple URL typesetting
\usepackage{booktabs}       % professional-quality tables
\usepackage{adjustbox}
\usepackage{amsfonts}       % blackboard math symbols
\usepackage{nicefrac}       % compact symbols for 1/2, etc.
\usepackage{microtype}      % microtypography
\usepackage{xcolor}         % colors
\usepackage[inline]{enumitem}

\usepackage{amsmath}
\usepackage{amssymb}
\usepackage{subfigure}
\usepackage{graphicx}
\usepackage{booktabs}
\usepackage{multirow}
\usepackage{multicol}
\usepackage{makecell}
\usepackage{colortbl}
\usepackage[redeflists]{IEEEtrantools}
\usepackage[english]{babel}
\usepackage{wrapfig}
\usepackage{graphicx}
\usepackage[labelfont=bf, skip=5pt]{caption}

\usepackage[ruled,vlined]{algorithm2e}

\usepackage{hyperref}
\usepackage[capitalise]{cleveref}

\usepackage{rotating}
\usepackage{soul}
\setul{0.8pt}{0.8pt}

\usepackage{wrapfig}

\newtheorem{proof}{Proof}
\newtheorem{theorem}{Theorem}

\newtheorem{corollary}{Corollary}

\renewcommand{\arraystretch}{0.87}
\makeatletter
\g@addto@macro\normalsize{%
  \abovedisplayskip 5pt plus 3pt minus 3pt%
  \belowdisplayskip \abovedisplayskip
  \abovedisplayshortskip 5pt plus 3pt minus 3pt%
  \belowdisplayshortskip 5pt plus 2pt minus 3pt%
}

\title{PDE+: Enhancing Generalization via PDE \\
with Adaptive Distributional Diffusion}

\author{%
\textbf{Yige Yuan}\textsuperscript{\rm 1,2}, 
\textbf{Bingbing Xu}\textsuperscript{\rm 1}\thanks{Corresponding author}, 
\textbf{Bo Lin}\textsuperscript{\rm 3}, \\
\textbf{Liang Hou}\textsuperscript{\rm 1,2}, 
\textbf{Fei Sun}\textsuperscript{\rm 1}, 
\textbf{Huawei Shen}\textsuperscript{\rm 1,2}\footnotemark[1], 
\textbf{Xueqi Cheng}\textsuperscript{\rm 1,2}\footnotemark[1]\\
\textsuperscript{\rm 1} CAS Key Laboratory of AI Security,\\ Institute of Computing Technology, Chinese Academy of Sciences, Beijing, China\\
\textsuperscript{\rm 2} University of Chinese Academy of Sciences\\
\textsuperscript{\rm 3} Department of Mathematics, National University of Singapore \\
\{yuanyige20z,xubingbing,houliang17z,sunfei,shenhuawei,cxq\}@ict.ac.cn, matbl@nus.edu.sg
}

\begin{document}

\maketitle
\begin{abstract}
The generalization of neural networks is a central challenge in machine learning, especially concerning the performance under distributions that differ from training ones. 
Current methods, mainly based on the data-driven paradigm such as data augmentation, adversarial training, and noise injection, may encounter limited generalization due to model non-smoothness. 
In this paper, we propose to investigate generalization from a Partial Differential Equation (PDE) perspective, aiming to enhance it directly through the underlying function of neural networks, rather than focusing on adjusting input data.
Specifically, we first establish the connection between neural network generalization and the smoothness of the solution to a specific PDE, namely ``transport equation''. 
Building upon this, we propose a general framework that introduces adaptive distributional diffusion into transport equation to enhance the smoothness of its solution, thereby improving generalization.
In the context of neural networks, we put this theoretical framework into practice as \textbf{PDE+} (\textbf{PDE} with \textbf{A}daptive \textbf{D}istributional \textbf{D}iffusion) which diffuses each sample into a distribution covering semantically similar inputs. This enables better coverage of potentially unobserved distributions in training, thus improving generalization beyond merely data-driven methods. 
The effectiveness of PDE+ is validated through extensive experimental settings, demonstrating its superior performance compared to SOTA methods.\footnote{Code is available: \url{https://github.com/yuanyige/pde-add}.}

\end{abstract}

\section{Introduction}

The generalization of neural networks is a fundamental challenge in the field of machine learning. 
It refers to the ability of neural networks to perform effectively under unobserved distributions, which may differ from those encountered during the training process~\cite{bousquet2002stability,lecun2015deep}. 
Pursuing superior generalization capability is essential as it ensures model adaptability to diverse real-world scenarios, guaranteeing reliable predictions and decisions.

\begin{figure}
    \centering
    \includegraphics[width=\linewidth]{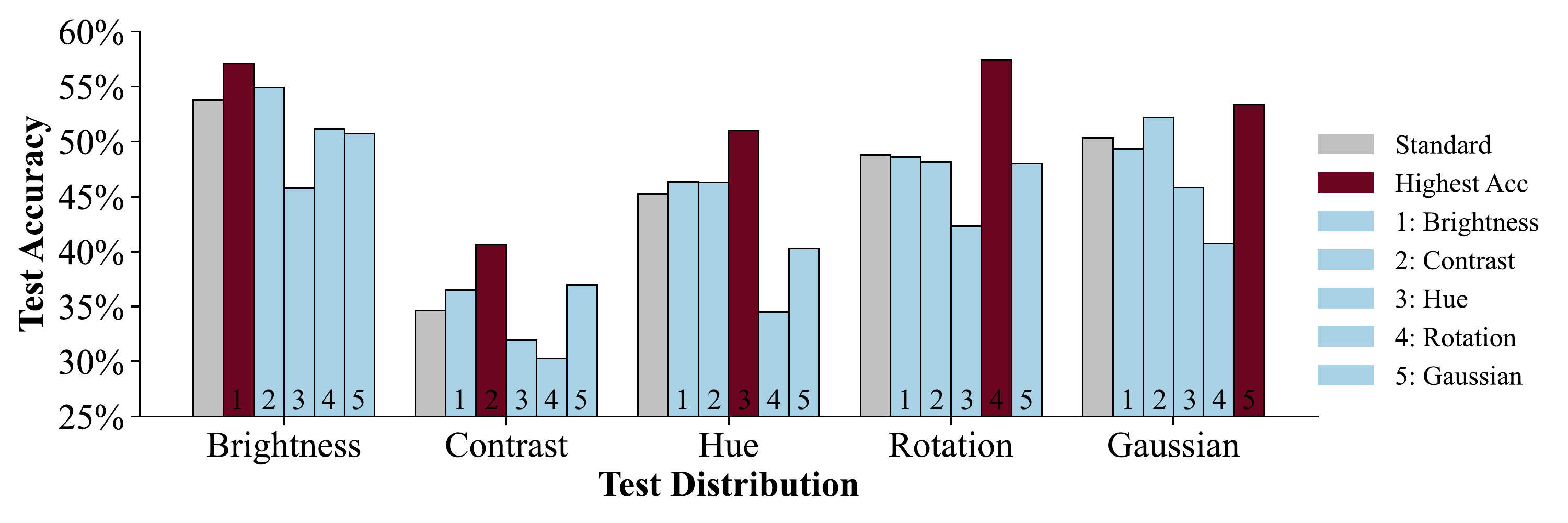}
    \caption{Model trained on six training distributions and evaluated on five corresponding test distributions. Model performs best on each test distribution is highlighted in red.}
    \label{Fig:motivation}
\end{figure}

Existing approaches for improving generalization mainly employ a data-driven paradigm~\cite{emmert2022taxonomy}, including 
data augmentation~\cite{DBLP:journals/jbd/ShortenK19}, adversarial training~\cite{madry2018towards}, and noise injection~\cite{DBLP:journals/neco/Bishop95}.
In terms of implementation, they primarily enhance the training samples via manipulating the original input~\cite{zhang2018mixup,hendrycks*2020augmix,madry2018towards,DBLP:journals/neco/An96} or transforming the hidden representations~\cite{verma2019manifold,lim2022noisy,Liu_2018_ECCV}.
However, such a data-driven paradigm usually cannot guarantee reliable generalization capabilities on unobserved distributions.
Taking data augmentation as an illustration, \cref{Fig:motivation} shows that the model can only achieve satisfactory generalization performance when the training data is subjected to augmentation similar to that of the testing data.
Analogous phenomena are also frequently observed in adversarial training and noise injection.
For instance, adversarial training can improve generalization on adversarial examples but often comes with the cost of performance on natural data~\cite{tsipras2018robustness,pmlr-v97-zhang19p}. 
Likewise, while injecting Gaussian noise can enhance generalization in the face of common corruptions, it risks $\sigma$-overfitting~\cite{kireev2022on}, i.e., overfitting to the particular Gaussian noise used in training.

The limited generalization capabilities of data-driven paradigm is due to model \emph{irregularity}~\cite{doi:10.1137/19M1265302}, i.e., the function learned by the neural network is non-smoothness. This may cause a problematic situation where semantically similar samples are encoded distantly, resulting in incorrect predictions. 
To address the irregularity issue, several approaches have been proposed to improve the smoothness of models~\cite{DBLP:journals/tsp/SokolicGSR17,than2021generalization,jin2020quantifying}, which helps to tackle the distribution shift problem~\cite{DBLP:conf/eccv/RodriguezLDL20}. 
Among them, Lipschitz continuity~\cite{DBLP:journals/tsp/SokolicGSR17,cisse2017parseval} enforces smoothness constraints on models through regularization or architectural restrictions, e.g., gradient regularization~\cite{165600} and spectral normalization~\cite{miyato2018spectral}. 
However, such restrictions often come at the cost of expressive power~\cite{anil2019sorting}.

In this paper, we go beyond the data-driven paradigm and propose to investigate generalization from a Partial Differential Equation (PDE)~\cite{DBLP:series/synthesis/2017Arrigo} perspective, aiming to directly introduce the smoothness constraint into the underlying function $\mathit{fn}$ of neural network, rather than manipulating input data.
The feasibility of this perspective is rooted in the intrinsic connection between neural networks and PDE~\cite{weinan2017proposal,DBLP:journals/corr/abs-1708-05115,DBLP:journals/corr/abs-1812-00174,doi:10.1073/pnas.1718942115}.
PDE describes a function that satisfies differential relationships, and neural networks can be regarded as a discrete numerical difference solver of PDE. 
That is to say, the underlying function of a neural network can be considered as the solution to PDE~\cite{DBLP:journals/corr/abs-1708-05115,DBLP:journals/corr/abs-1812-00174,doi:10.1073/pnas.1718942115}.
From such perspective, we can leverage the vast prior knowledge of PDE to constrain the underlying function of neural network, thus encouraging the resulting neural networks to exhibit specific desired properties, e.g., smoothness~\cite{doi:10.1137/19M1265302}, well-posedness~\cite{haber2017stable}, and hyperbolicity~\cite{DBLP:conf/nips/EliasofHT21}.

The above fundamental connection inspires us to establish the connection between neural network generalization and the smoothness of PDE solution. 
Specifically, we initially model the neural network as the solution of a specific type of PDE, referred to as transport equation (TE)~\cite{DBLP:journals/corr/abs-1708-05115}, which is often employed to describe the transportation of a quantity within a space. 
Then, a diffusion term is introduced into the TE, which has been proven to smooth the solution~\cite{doi:10.1137/19M1265302,ladyzhenskaya1968linear}. The core of such paradigm is this key question: \emph{What type of diffusion term is appropriate for a neural network to achieve effective generalization?} 
To answer it, we propose a general framework that introduces adaptive distributional diffusion into transport equation to enhance the smoothness of its solution. 
Such diffusion ensures suitable smoothness by treating the diffusion scope of each sample as a distribution that should cover the potential semantically similar inputs, thus improving generalization.

In the context of neural networks, we put this theoretical framework into practice as \textbf{PDE+} (\textbf{PDE} with \textbf{A}daptive \textbf{D}istributional \textbf{D}iffusion, PDE-ADD) to achieve generalization.
Specifically, we introduce adaptive distributional diffusion into the neural network, which performs diffusion centered on each data point.
The scope of each diffusion is modeled as a distribution, determined adaptively by multiple augmentations of the input. This enables better coverage of potentially unobserved distributions and improves generalization beyond data-driven approaches.
The effectiveness of PDE+ is validated on various distributions, including clean samples and various common corruptions. 
The consistent improvements demonstrate the superior performance of our method over state-of-the-art methods. 

\newpage
\textbf{Our main contributions include:} 

(1) \textit{A promising paradigm}:  we investigate generalization from a Partial Differential Equation (PDE) perspective. To the best of our knowledge, we are the first to achieve generalization by establishing connections between the generalization of neural networks and the smoothness of TE solutions.

(2) \textit{An innovative method}: we propose an adaptive distributional diffusion term to incorporate smoothness into a neural network and instantiate it as PDE+, enabling better coverage of potentially unobserved distributions in training and improves generalization compared to data-driven methods.

(3) \textit{Solid experiments}: extensive experiments reveal  PDE+ outperforms baselines across unobserved distributions, e.g., the improvements are up to 3.8\% in Acc and 7.7\% in mCE.

\section{Related Work} 

In this section, we briefly review two lines of research that close to our work: the generalization of neural networks and differential equations based neural networks. 
Detailed introduction of related works can be found in \cref{App:Related_Work}.

\paragraph{Generalization of Neural Networks.}
Current data-driven methods encompass data augmentation, adversarial training, and noise injection. 
Data augmentation is a widely adopted technique to enhance generalization, employing various strategies such as Mixup~\cite{zhang2018mixup}and AugMix~\cite{hendrycks*2020augmix}. 
Adversarial training is a robust optimization approach for improving adversarial generalization~\cite{DBLP:journals/corr/GoodfellowSS14} while potentially compromising non-adversarial generalization~\cite{tsipras2018robustness,pmlr-v97-zhang19p}. Notable works in this area include PGD~\cite{madry2018towards}, TRADES~\cite{pmlr-v97-zhang19p}, and RLAT~\cite{kireev2022on}. 
Noise injection  introduces noise into input data~\cite{an1996effects}, activations ~\cite{pmlr-v48-gulcehre16}, or hidden layers~\cite{NEURIPS2020_c16a5320}, whose noise magnitude can be sensitive and susceptible to overfitting~\cite{kireev2022on}.  
Lipschitz continuity is often used to ensure model generalization~\cite{165600,miyato2018spectral,liu2023decoupled}, but its strict constraint can restrict a model's capabilities~\cite{anil2019sorting}.
Our method diverges from above approaches, as we directly constrain the smoothness of the neural network's underlying function rather than fitting a finite set of input data like data-driven methods. Although our method shares the concept of smoothness with Lipschitz, 
it avoids compromising the model's capabilities.

\paragraph{Differential Equations based Neural Networks}
The connection between continuous dynamical systems and residual neural networks~\cite{he2016deep} is initially established in~\cite{weinan2017proposal}. 
Subsequently, numerous studies have delved into the relationships between various neural network architectures and different types of differential equations~\cite{DBLP:conf/iclr/LuZLD18, DBLP:journals/corr/abs-1708-05115, DBLP:journals/corr/abs-1812-00174}.
Since then, researchers have started to explore the beneficial properties of differential equations to enhance neural networks~\cite{661124,10.5555/3524938.3525483,doi:10.1137/19M1265302}.

\section{Generalization under PDEs with Adaptive Distributional Diffusion}

This section introduces the theoretical motivation and framework behind our method. We begin by establishing connections between PDEs and neural networks, thereby transforming the generalization of neural networks into the smoothness of PDE solutions. 
Our innovative adaptive distributional diffusion term is then introduced to enhance the smoothness of solutions, which improves generalizability.

\subsection{Neural Network as the Solution of Transport Equation}
\label{Sec3:Transport_Equation_Modeling_for Neural_Networks}

Partial Differential Equation (PDE)~\cite{DBLP:series/synthesis/2017Arrigo} is an equation containing an unknown function $u$ of multiple variables and its partial derivatives.
The connection between PDEs and neural networks has been discussed in~\cite{weinan2017proposal}, where neural networks could be interpreted as a numerical scheme to solve PDEs. Such connection allows us to take advantage of PDE, such as the properties of solution as well as the numerical schemes, to obtain a better neural network. In this section, we make use of the transport equation (TE), which is one special form of PDE, to interpret neural networks.

TE describes the concentration of a quantity transport in a fluid~\cite{pogodaev2016optimal,munson2006fundamentals} (\cref{Eq:TE}), which is suitable to model the feature transformation of data flow.
This observation has also been discussed in~\cite{DBLP:journals/corr/abs-1708-05115,DBLP:journals/corr/abs-1812-00174}

\begin{equation}
\frac{\partial u}{\partial t}(\mathbf{x}, t)+F(\mathbf{x}, \boldsymbol{\theta}(t)) \cdot \nabla u(\mathbf{x}, t)=0
\label{Eq:TE}
\end{equation}
where $u(\mathbf{x},t)$ denotes a function of concentration, which can be viewed as the underlying function of a neural network.
$t \in (0,1)$ denotes time, serving as the continuation of network layers. 
$\mathbf{x} \in \mathbb{R}^d$ denotes a variable in space, serving as the variable for data representation in terms of neural networks.
$\nabla$ represents gradient, and $F(\mathbf{x}, \boldsymbol{\theta}(t))$ is the velocity field, serving as the continuation for network structures and parameters. In terms of neural networks, the changing of representation through layers can be viewed as a transport process over time. The representation is transported through each layer, where the parameters of each layer serve as a velocity field aiming to make changes to the sample representations and transport it to the next layer. 
Given the parameters of all layers,
the representation transforms from the original input to final output, acting like a transport of data flow as illustrated in the top subfigure of \cref{Fig:PDEADD}.

$u(\mathbf{x},t)$ represents the value obtained by transporting the variable $\mathbf{x}$ through a series of $F(\mathbf{x}, \boldsymbol{\theta}(t))$ from time $t$ until the terminal.
The terminal condition of TE is enforced at $t = 1$ as $u(\mathbf{x},1)=o(\mathbf{x})$, where $o(\mathbf{x})$ denotes the output function such as softmax~\cite{gold1996softmax}.
Let $\hat{\mathbf{x}}$ denote the input feature. The original data-label pair $(\hat{\mathbf{x}},y)$ is given at $t = 0$, and an optimal network $u^{*}$  should exactly maps $\hat{\mathbf{x}}$ to $y$, 
i.e., $u^{*}\left(\hat{\mathbf{x}},0\right)=y$.
Obtaining the network is equivalent to solving the numerical solution of TE at $t=0$ as $u(\hat{\mathbf{x}},0)$, where the method of characteristics~\cite{sarra2003method} can be effectively employed. 
The main idea of the characteristics is to solve PDE via an ordinary differential equation (ODE) defining the characteristic curves of original PDE, which is shown in \cref{Eq:ODE}. 
Then the solution of PDE can be acquired by following these curves in \cref{Eq:U_ODE}.
\begin{align}
\mathrm{d} \mathbf{x}(t) &= F(\mathbf{x}(t), \boldsymbol{\theta}(t))\, \mathrm{d} t \label{Eq:ODE}\\
u(\hat{\mathbf{x}}, 0) &= o\Big(\hat{\mathbf{x}} + \int_0^1 F(\mathbf{x}(t),\boldsymbol{\theta}(t)) \mathrm{d} t\Big)  \label{Eq:U_ODE}
\end{align}

To solve \cref{Eq:ODE} numerically, we adopt Euler method~\cite[Chapter 2]{butcher2003numerical} as shown in \cref{Eq:ODE_Euler}, which recovers the formulation of ResNet~\cite{he2016deep}.
$l {\in} \{1,\dots, L\}$ is the network layer index, serving as a discrete slicing to continuous time $t$.
$\mathbf{h}_l$ and $\boldsymbol{\theta}_l$ are representations and parameters at layer $l$, respectively.
\begin{align}
\mathbf{h}_{l+1} &= f(\mathbf{h}_l,\boldsymbol{\theta}_l)+\mathbf{h}_l \label{Eq:ODE_Euler} \\
u(\hat{\mathbf{x}}, 0) &=o\Big(\hat{\mathbf{x}} + \sum_{l=1}^L f(\mathbf{h}_l,\boldsymbol{\theta}_l)\Big) \label{Eq:U_ODE_Euler} 
\end{align}
Overall, neural network, particularly ResNet can be seen as a solution to TE. This connection lays a solid foundation to achieve desired properties of neural networks by constraining the solution of TE.

\subsection{Improving Generalization via Enhancing  the Smoothness of TE Solution}

Smoothness has been demonstrated to be strongly linked to generalization, as it facilitates models to generalize beyond the training distribution~\cite{pmlr-v137-rosca20a,DBLP:conf/eccv/RodriguezLDL20}, enhances model robustness against small perturbations~\cite{cisse2017parseval,DBLP:journals/tsp/SokolicGSR17}, and plays a significant role in generalization quantization~\cite{jin2020quantifying,DBLP:journals/corr/abs-2207-02093} as well as uncertainty estimation~\cite{van2020uncertainty,liu2020simple}. 
Building upon the insights, we propose to achieve generalization from the perspective of PDEs by modeling neural networks as solutions to PDEs and transforming the generalization goal of neural networks into smoothness goal of a solution to PDEs. 

To enhance the smoothness of solution $u(\mathbf{x},t)$, we leverage knowledge from PDE field to introduce a diffusion term~\cite{ladyzhenskaya1968linear} $\Delta u(\mathbf{x}, t)$ into TE as \cref{Eq:TE_FixDiff}. 
The diffusion term corresponds to the Laplacian, i.e., the second-order derivative with respect to $\mathbf{x} \in \mathbb{R}^d$, as illustrated in \cref{Eq:Diffusion_Term}. 
Here, $\Delta$ denotes the Laplacian operator, and $\sigma \neq 0$ is a coefficient for the diffusive magnitude.
\begin{align}
&\small \frac{\partial u}{\partial t}(\mathbf{x}, t)\!+F(\mathbf{x}, \boldsymbol{\theta}(t))\! \cdot \nabla u(\mathbf{x}, t) \!+ \frac{1}{2}\sigma^2 \!\cdot \Delta u(\mathbf{x}, t)=0
\label{Eq:TE_FixDiff}\\
&\Delta u =  {\partial}^2 u/\partial x_1^2 + \partial^2 u/\partial x_2^2 + \dots + \partial^2 u /\partial x_d^2
\label{Eq:Diffusion_Term}
\end{align}

\begin{theorem}[Proved in \cref{App:Proof_Th1}]
\label{Them:TEGen}
Given TE with diffusion term (\cref{Eq:TE_FixDiff}) with terminal condition $u(\mathbf{x}, 1) = o(\mathbf{x})$, where $F(\mathbf{x}, \boldsymbol{\theta}(t))$ be a Lipschitz function in both $\mathbf{x}$ and $t$, $o(\mathbf{x})$ be a bounded function. Then, for any small $\delta$,
$|u(\mathbf{x}+\delta, 0)-u(\mathbf{x}, 0)| \leq C\big(\frac{\|\delta\|_2}{\sigma}\big)^\alpha$ holds for constant $\alpha>0$ if $\sigma \leq 1$, where
$\|\delta\|_2$ is the $\ell_2$ norm of $\delta$, and $C$ is a constant that depends on $d,\|o\|_{\infty}$, and $\|F\|_{L_{\mathbf{x}, t}^{\infty}}$.
\end{theorem}

\begin{corollary}[Proved in \cref{App:Proof_Co1}]
\label{Coro:TEGen}

Generalization Error (GE) of model $u(\mathbf{x},0)$ trained on training set $s_N$ is upper bounded by diffusion $\sigma$. For any $\epsilon > 0$, the following inequality holds with probability at least $1-\epsilon$.
For more details about the notations used, please refer to \cref{App:Proof_Co1}.
\begin{equation}
\footnotesize \mathrm{GE}\left(u(\mathbf{x},0), s_N \right) \leq C \cdot L \left(\frac{\|\delta'\|_2}{\sigma}\right)^\alpha \! +M \sqrt{ \frac{2K\ln 2+2 \ln (1 / \epsilon)}{N}}
\end{equation}
\end{corollary}

\begin{figure}[!t]
    \centering
    \includegraphics[width=1.0\linewidth]{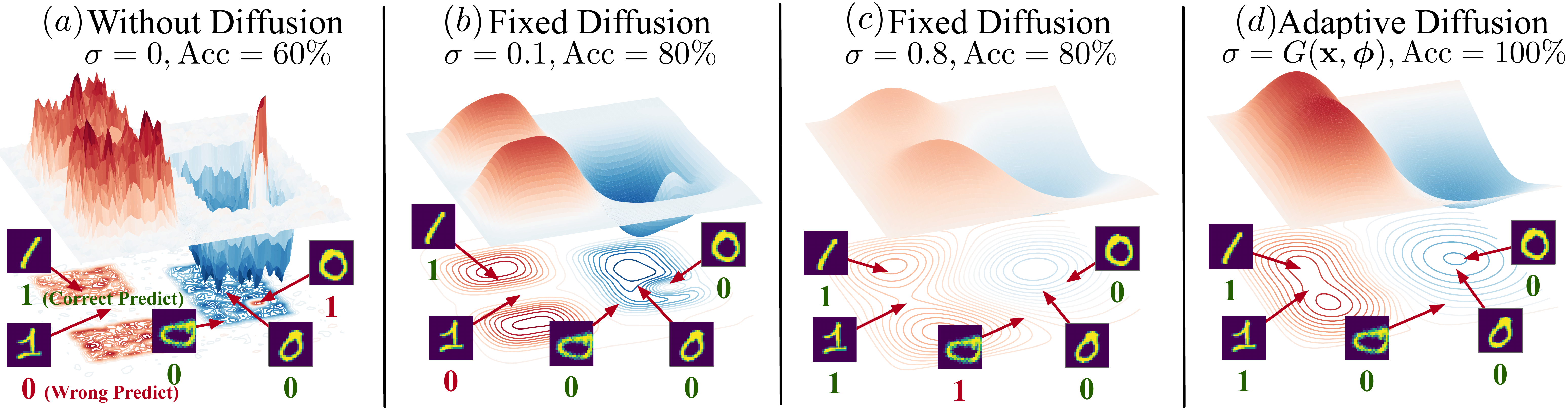}
    \caption{Solutions to 2D TE differs in the diffusion $\sigma$. The upper displays function surface, the lower exhibits its contour, with samples showing its true and predicted label.}
    \label{Fig:Smoothness}
\end{figure}

Typically, $\sigma$ is chosen as a fixed scalar, imposing an uniform diffusion scale across entire data space~\cite{doi:10.1137/19M1265302}. Fixed diffusion brings smoothness into TE solution, but it neglects structure of solution for different $\mathbf{x}$.
It cannot achieve an optimal diffusion scale for network across  data space, as different locations require diverse diffusion scales based on their distance to other samples or class boundaries. 
To intuitively introduce the influence of diffusion, we illustrate the solution surface of 2D transport equation under different diffusion terms in \cref{Fig:Smoothness}. 
No diffusion in (a) results in highly irregular surface.
Fixed diffusion with a small coefficient in (b) imposes insufficient smoothness for same class. Larger coefficient in (c) imposes 
over-smoothness for different classes.
It reveals that a fixed coefficient can result in over-smoothness which diminishes variability, or insufficient smoothing. Thus, a new diffusion term is required to improve generalizability.

\subsection{Adaptive Distributional Diffusion for Generalization} \label{Sec3:Adaptive_Distributional_Diffusion_for_Generalization}

With concerns draw above, a crucial question arises:
\begin{center}

\textit{What type of diffusion term is appropriate for a neural network to achieve effective generalization?}
\end{center}

To address it, we claim that a good diffusion term for generalization should satisfy two goals: ``Adaptive'' and ``Distributional''. 
``Adaptive'' stands for that the diffusion varies in magnitude for every point across the entire space. 
``Distributional'' treats the diffusion scope of each point as a distribution. For any input from the data space at any time step, the distribution should only encompass the inputs that are potentially similar to the central point in semantics.
This mechanism allows for better coverage of potential unseen distributions and improved generalization compared to data-driven methods.

To achieve the above goals, we propose an \textbf{Adaptive Distributional Diffusion (ADD)} term and introduce it into TE as presented in \cref{Eq:TE_AdaDiff}. 
Rather than using a fixed scalar, our term incorporates a coefficient function $G(\mathbf{x}, \boldsymbol{\phi}(t))$ that takes  sample $\mathbf{x}$ as input and outputs its diffusion scale, 
exhibiting different diffusion properties,
based on the parameters $\boldsymbol{\phi}$ at each time step $t$.
The benefits of the term can be illustrated in \cref{Fig:Smoothness}(d), which allows for different smoothing effects across space in accordance with the principle of ``adaptive''. Meanwhile, data spaces with similar semantics or within the same class can achieve smoothness in their scope, and those within different classes can avoid over-smoothness and maintain discrepancy. These satisfy the principle of ``distributional''.
\begin{equation}
\frac{\partial u}{\partial t}(\mathbf{x}, t) + F(\mathbf{x}, \boldsymbol{\theta}(t)) \cdot \nabla u(\mathbf{x}, t)+  \frac{1}{2} G(\mathbf{x}, \boldsymbol{\phi}(t))^2 \cdot \Delta u(\mathbf{x}, t)=0
\label{Eq:TE_AdaDiff}
\end{equation}

\subsection{Deriving Neural Network from Transport Equation with ADD}

Introducing adaptive distributional diffusion into TE as \cref{Eq:TE_AdaDiff} can realize the smoothness of the solution of TE, and thus encourage the resulting neural networks to exhibit generalization. In the following, we solve TE with ADD (\cref{Eq:TE_AdaDiff}) to derive its corresponding neural network.

\begin{theorem}[Proved in \cref{App:Proof_Th2}]
\label{Them:Feynman-Kac}
    TE with adaptive distributional diffusion term (\cref{Eq:TE_AdaDiff})can be solved using the Feynman-Kac formula~\cite{10.2307/1990512}, The result is shown in \cref{Eq:FK_U,Eq:FK_SDE}, where $B_t$ represents the Brownian motion~\cite{PhysRev.36.823}. 
\begin{align}
u(\hat{\mathbf{x}}, 0) &= \mathbb{E}\left[o(\mathbf{x}(1)) \mid \mathbf{x}(0)=\hat{\mathbf{x}}\right]   \label{Eq:FK_U}\\
\mathrm{d} \mathbf{x}(t) &= F(\mathbf{x}(t), \boldsymbol{\theta}(t))\, \mathrm{d} t+ G(\mathbf{x}(t), \boldsymbol{\phi}(t)) \cdot \mathrm{d} B_t \label{Eq:FK_SDE}
\end{align}
\end{theorem}

The result is a conditional expectation with respect to the initial value problem of stochastic differential equation (SDE, \cite{kloeden1992stochastic}) in \cref{Eq:FK_SDE}. 
To obtain the final functional form of our neural network, we adopt the Euler–Maruyama method~\cite{DBLP:journals/siamrev/GelbrichR95} to compute the solution of SDE numerically as follows.
\begin{equation}
\begin{split}
u(\hat{\mathbf{x}}, 0) &= \mathbb{E}\left[o(\mathbf{h}_L) \mid \mathbf{h}_0=\hat{\mathbf{x}}\right] \\
\mathbf{h}_{l+1} &=\mathbf{h}_l + f\left(\mathbf{h}_l,\boldsymbol{\theta}_l\right)+ g(\mathbf{h}_l, \boldsymbol{\phi}_l) \cdot \mathcal{N}(\mathbf{0}, \mathbf{I})
\label{Eq:SDE_Euler}
\end{split}
\end{equation}

\section{PDE+~:~An Neural Network Instantiation}

\begin{figure*}[!t]
    \centering
    \includegraphics[width=1.0\linewidth]{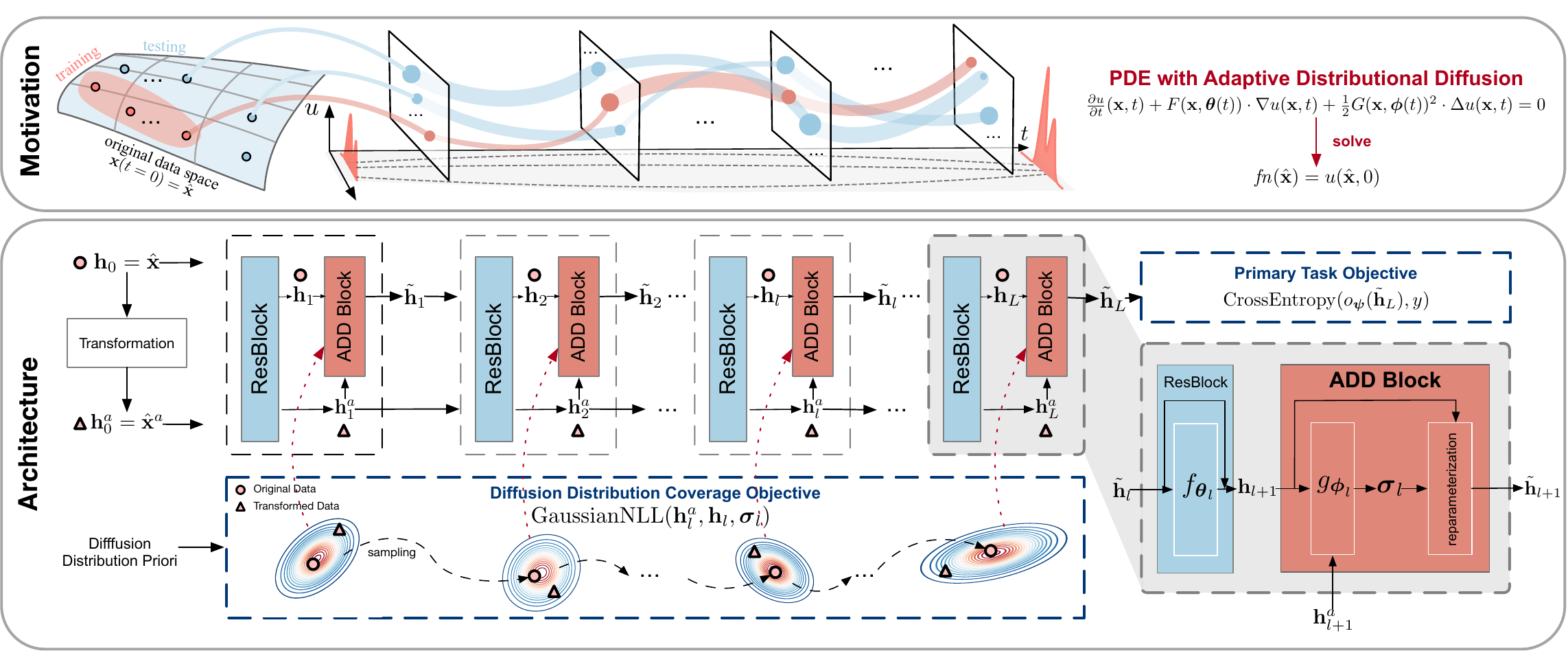}
    \caption{Motivation and architecture of PDE+, the upper illustrates our motivation of solving transport equation with adaptive distributional diffusion (ADD) to derive the functional form of neural network. The lower is the neural network instantiation, which comprises a series of blocks that contain a residual block followed by an ADD block. The architecture of the ADD block is enclosed grey frame on the right. The learning objectives are enclosed in two blue frames.}
    \label{Fig:PDEADD}
\end{figure*}

This section is for the instantiation of our framework \textbf{PDE+}: \textbf{PDE} with \textbf{A}daptive \textbf{D}istributional \textbf{D}iffusion (PDE-ADD).

\subsection{Overall Architecture} %
PDE+ is a neural network instantiation of PDE solution formulated in \cref{Eq:SDE_Euler}, where $\mathbf{h}_{l+1} =\mathbf{h}_l + f\left(\mathbf{h}_l,\boldsymbol{\theta}_l\right)$ is the  formulation for residual block, and $g(\mathbf{h}_l, \boldsymbol{\phi}_l) \cdot \mathcal{N}(\mathbf{0}, \mathbf{I})$ is implemented as our adaptive distributional diffusion block, dubbed as ADD block.  
As shown in \cref{Fig:PDEADD}, the residual block is denoted as $f_{\boldsymbol{\theta}_l}$ parameterized by $\boldsymbol{\theta}_l$, where $l \in \{1,\dots,L\}$ denotes the block index. 
ADD block is denoted as $g_{\boldsymbol{\phi}_l}$ parameterized by $\boldsymbol{\phi}_l$. The overall architecture of PDE+, denoted as $\mathit{fn}_{\boldsymbol{\theta},\boldsymbol{\phi}}$
is the composition of $L$ blocks, where each block contains a residual block followed by our ADD block.

\subsection{Adaptive Distributional Diffusion Block}
\label{Sec4:Adaptive_Distributional_Diffusion_Block}
\cref{Fig:PDEADD} illustrates the structure of ADD block, which takes the output from residual block $\mathbf{h}_{l}$ as input, and outputs the scale $\boldsymbol{\sigma}_l$ for diffusion (\cref{Eq:PDE+_var}). 
Then a reparameterization trick~\cite{kingma2013auto} of $\mathbf{h}_{l}$ and $\boldsymbol{\sigma}_l$ under the prior of Gaussian distribution is conducted to obtain the final output $\mathbf{\tilde{h}}_l$ (\cref{Eq:PDE+_h}). 
\begin{align}
\boldsymbol{\sigma}_l =\, & g_{\boldsymbol{\phi}_l}(\mathbf{h}_{l}) \label{Eq:PDE+_var}\\
\mathbf{\tilde{h}}_l =\, & \mathbf{h}_{l} + \boldsymbol{\sigma}_l \cdot \mathcal{N}(\mathbf{0}, \mathbf{I})  \label{Eq:PDE+_h}
\end{align}

As introduced in \cref{Sec3:Adaptive_Distributional_Diffusion_for_Generalization}, the principle of ADD blocks is ``adaptive'' and ``distributional''. ``Adaptive'' is implemented by replacing the fixed diffusion with the learnable $\boldsymbol{\sigma}_l$. ``Distributional'' means that for any input from the data space at any given time step, the diffusion scope should encompass the potential neighbors that exhibit semantic similarity. 
To achieve this, semantically similar samples are utilized as guidance. 
Define training dataset $s_N$ containing $N$ training samples of $C$ classes $s_N=\{(\mathbf{x}_n,y_n)\mid n \in 1,2 \dots N\}$.
Let $\mathbf{x}_n^a$ represent samples that share semantic similarity with $\mathbf{x}_n$, such as augmented samples, style-transferred samples, or adversarial attack samples. We hope the diffusion distribution scope of $\mathbf{x}_n$ can cover $\mathbf{x}_n^a$.

To achieve this, we let the original samples pass through the whole block with both residual block and ADD block, and the semantically similar samples only go through residual block without diffusion. Denote $I$ as the identity function where $I(x)=x$. The $l$-th layer's representation of original samples and their semantically similar counterparts can be formulated as... \cref{Eq:Train_h,Eq:Train_ha}, . 
\begin{align}
\tilde{\mathbf{h}}_l &= (g_{\boldsymbol{\phi}_l} \circ (f_{\boldsymbol{\theta}_{l-1}}+I) \circ \cdots \circ g_{\boldsymbol{\phi}_2}\circ (f_{\boldsymbol{\theta}_1}+I))(\mathbf{x})    \label{Eq:Train_h}\\
\mathbf{h}_l^a &= ((f_{\boldsymbol{\theta}_{l-1}}+I) \circ \cdots \circ (f_{\boldsymbol{\theta}_1}+I)) (\mathbf{x^a})  \label{Eq:Train_ha}
\end{align}

For every block, the diffused hidden representation $\mathbf{\tilde{h}}_l$ can be regarded as a sampling from a Gaussian distribution $\mathcal{N}(\mathbf{h}_l, \boldsymbol{\sigma}_l)$, where the representations of semantically similar samples $\mathbf{h}_l^a $ should be covered. This objective can be implement via maximizing the probability of $\mathbf{h}_l^a $ under $\mathcal{N}(\mathbf{h}_l, \boldsymbol{\sigma}_l)$ denoted as $p_{\boldsymbol{\phi}}(\mathbf{h}_l^a \mid \mathbf{h}_l)$, which is equivalent to minimizing its negative log-likelihood. We named such objective as \textit{diffusion distribution coverage objective} shown in \cref{Eq:Obj_nll}, guiding only the parameters of diffusion blocks $\boldsymbol{\phi}$. 
\begin{align}
&\min_{\boldsymbol{\phi}} \underset{\mathbf{x} \sim s_N}{\mathbb{E}} -\sum_{l=1}^L \log  p_{\boldsymbol{\phi}_l}(\mathbf{h}_l^a \mid \mathbf{h}_l) = -\frac{1}{2N} \sum_{n=1}^N \sum_{l=1}^L  \left[ \log g_{\boldsymbol{\phi}_l}(\mathbf{h}_{l}) + \frac{(\mathbf{h}_{n,l}^a-\mathbf{h}_{n,l})^{2}}{g_{\boldsymbol{\phi}_l}(\mathbf{h}_l)}\right]
\label{Eq:Obj_nll}
\end{align}

From a distributional point of view, the intuitive interpretation of our adaptive distributional diffusion is treating each sample as one distribution whose scope includes its semantic similar samples. Under such view, the basic residual block without diffusion treats each sample as a Dirac distribution~\cite{COHEN1991312} and our ADD block transforms it into Gaussian distribution. To broaden the distribution and enhance generalization, we advance from a single Gaussian to a Gaussian mixture~\cite{reynolds2009gaussian}, as it is a universal approximator of densities~\cite{goodfellow2016deep}. 
Notably , we do not model the Gaussian mixture distribution directly. Rather, we allow both the original sample and its augmentations to diffuse simultaneously, effectively acting as different Gaussian centers. 
As a result, the superimposition of these multiple single Gaussians manifests as a mixed Gaussian from a macroscopic perspective. This implementation can be easily achieved in one line of code, as shown in \cref{Alg:Train}(\cref{Alg:Concat}) from \cref{App:Algorithm}.

\subsection{Learning Objectives}

PDE+ consists of two learning objectives: a diffusion distribution coverage objective for every ADD block (\cref{Eq:Obj_nll}) and a primary task objective for the entire network. 
The primary task objective ensures the correctness of learning representations under diffusion. 
Define the output of $\mathbf{x}_n$ throughout the whole model $\mathit{fn}$  as $\mathbf{\tilde{h}}_{n,L} =\mathit{fn}_{\boldsymbol{\theta},\boldsymbol{\phi}}(\mathbf{x}_n)$.
The primary task objective is shown in \cref{Eq:Obj_cla}, where $o_{\boldsymbol{\psi}}$ stands for output layer parameterized by $\boldsymbol{\psi}$. The samples diffused throughout $\mathit{fn}$ to obtain a classification probability via softmax, guiding the learning of all parameters, including residual blocks $\boldsymbol{\theta}$, diffusion blocks $\boldsymbol{\phi}$ and output $\boldsymbol{\psi}$ via cross-entropy. The algorithmic pseudocode for both training and testing phase can be found in \cref{Alg:Train,Alg:Test} in \cref{App:Algorithm}. 
\begin{align}
    &\min_{\boldsymbol{\theta}, \boldsymbol{\phi}, \boldsymbol{\psi}} \underset{(\mathbf{x},y) \sim s_N}{\mathbb{E}} -\log p_{\boldsymbol{\theta}, \boldsymbol{\phi}, \boldsymbol{\psi}} (y \mid \mathbf{x}) = -\frac{1}{N} \sum_{n=1}^N \left[\log \frac{\exp(o_{\boldsymbol{\psi}}(\mathbf{\tilde{h}}_{n,L})_{y_n})}{\sum_{c=1}^C \exp(o_{\boldsymbol{\psi}}(\mathbf{\tilde{h}}_{n,L})_c)} \right]_{y_n}
\label{Eq:Obj_cla}
\end{align}

\section{Experiments}

\begin{table}[!t]
\renewcommand{\arraystretch}{1.02}
\caption{Comparisons of PDE+ and baselines on CIFAR-10(C), CIFAR-100(C) and Tiny-ImageNet(C) based on ResNet-18. The corruption is evaluated under all severity level and the severest level. The best result is highlighted in \textbf{boldface}.The abbreviations means Standard (Std), Lipschitz (Lip), Noise Injection (NI), Data Augmentation (DA), Adversarial Training (AT).}
\centering
\begin{adjustbox}{width=\textwidth}
\setlength{\tabcolsep}{0.5mm}
\begin{tabular}{llccccc ccccc ccccc}
\toprule
& \multirow{4}{*}{Method} & \multicolumn{5}{c}{CIFAR-10(C)} & \multicolumn{5}{c}{CIFAR-100(C)} & \multicolumn{5}{c}{Tiny-ImageNet(C)} \\
\cmidrule(lr){3-7} 
\cmidrule(lr){8-12}
\cmidrule(lr){13-17}
& & Clean &  \multicolumn{2}{c}{Corr Severity All} & \multicolumn{2}{c}{Corr Severity 5} & Clean &  \multicolumn{2}{c}{Corr Severity All} & \multicolumn{2}{c}{Corr Severity 5} & Clean &  \multicolumn{2}{c}{Corr Severity All} & \multicolumn{2}{c}{Corr Severity 5} \\
\cmidrule(lr){3-3} \cmidrule(lr){4-5} \cmidrule(lr){6-7} 
\cmidrule(lr){8-8} \cmidrule(lr){9-10} \cmidrule(lr){11-12}
\cmidrule(lr){13-13} \cmidrule(lr){14-15} \cmidrule(lr){16-17}
& & Acc ($\uparrow$)  & Acc ($\uparrow$) & mCE ($\downarrow$) & Acc ($\uparrow$) & mCE ($\downarrow$) & Acc ($\uparrow$)  & Acc ($\uparrow$) & mCE ($\downarrow$)& Acc ($\uparrow$) & mCE ($\downarrow$) & Acc ($\uparrow$)  & Acc ($\uparrow$) & mCE ($\downarrow$)& Acc ($\uparrow$) & mCE ($\downarrow$)\\
\midrule
Std & ERM & 95.35 & 74.63 & 100.00  & 57.19 & 100.00 & 77.71 & 49.27 & 100.00  & 33.18 & 100.00 & 54.02 & 25.57& 100.00& 15.54 &100.00\\
\midrule
Lip & GradReg & 93.64 & 77.62 & 96.29 & 62.33 & 91.52 & 73.80 & 52.16 & 96.95 & 37.33 & 94.49 &52.01&29.20& 95.13&19.91&94.86\\
\midrule
\multirow{3}{*}{NI} 
& EnResNet & 83.33 & 74.34 & 137.98& 66.87 & 63.72 & 67.11 & 49.28 & 103.61 & 40.24 & 83.56 &49.26&25.83&100.18&19.01&96.55\\
& RSE & 95.59 & 77.86 & 94.12 &  63.66 & 89.08 & 77.98 & 53.73 & 94.10 & 38.03 & 92.88 & 53.74 &27.99&96.81&18.92 & 96.11\\
& NFM* & 95.40 & 83.30 & - &- &- & 79.40 & 59.70 & -&- &- &- &- &- &- &- \\
\midrule
\multirow{5}*{DA} 
& Gaussian & 92.50 & 80.46
 & 100.03 & 68.08 & 87.22  & 71.87 & 54.24 & 98.34 & 41.77 & 89.81 & 48.89 & 32.92 & 90.48& 24.57 & 89.56 \\
& Mixup* & \textbf{95.80} & 80.40 & -& - &- & \textbf{79.70} & 54.20 & -&-&-&-&-&-&-&-\\
& DeepAug* & 94.10 & 85.33 & 64.63 & 77.29 & 60.05 &-&-&-&-&-&\bf 54.90&-&-&-&-\\
& AutoAug & 95.61 & 85.37 & 61.74  & 75.12 & 62.07  & 76.34 & 58.72 & 83.12 & 45.38 & 82.84 & 52.63 &35.14&87.67&25.36&88.54\\
& AugMix & 95.26 & 86.24 & 60.44 & 76.06 & 59.96  & 77.11 & 61.93 & 77.51   & 48.99 & 77.52 &52.82&37.74&84.06&28.66& 84.69\\
\midrule
\multirow{4}*{AT} 
& PGD$_{\ell_{\infty}}$ & 93.52 & 82.17 & 86.53 & 70.10 & 78.20 & 71.78 & 55.03 & 93.49  & 42.04 & 88.17 &49.94&32.54&90.65&23.47&90.63 \\
& PGD$_{\ell_2}$ & 93.91 & 83.07 & 81.06 & 70.97 & 75.17 & 72.50  & 56.09 & 91.65  & 42.82 & 87.33 &51.08&33.46&89.37 &24.00&89.92\\
& RLAT & 93.23 & 83.67 & 80.98 & 72.73 & 72.59 & 71.10 &  56.54 & 91.98 & 44.27 & 86.24 &50.24&33.13&89.83&24.46&89.47 \\
& RLAT$_{\text{Augmix}}$ & 94.73 & 88.28 & 55.60 & 80.37 & 51.56  & 75.06 & 62.77 & 77.38 & 51.60 & 74.24 & 51.29 &  37.92&83.69&29.05&84.17 \\
\midrule
\rowcolor{gray!20}
Ours & PDE+  & 95.59 & \textbf{89.11} & \textbf{48.07} & \textbf{82.81} & \textbf{44.97}  & 78.84 & \textbf{65.62} & \textbf{69.68} & \textbf{54.22} & \textbf{69.43} &53.72&\bf 39.41&\bf 81.80&\bf 30.32&\bf 82.68\\ 
\bottomrule
\end{tabular}

\end{adjustbox}
\label{Tab:cifar_ood}
\end{table}

In this section, we empirically evaluates PDE+ through the following questions. Due to the space limitations, more comprehensive experiments including full results on corruptions and diffusion scale analysis are provided in \cref{App:Experiments}.
\begin{itemize}
    \item (Q1) Does PDE+ improve generalization compared to SOTA methods on various benchmarks?
    \item (Q2) Does PDE+ learns appropriate diffusion distribution coverage?
    \item (Q3) Does PDE+ improve generalization beyond observed (training) distributions?
\end{itemize}

\begin{wraptable}{l}{8cm}
\renewcommand{\arraystretch}{0.3}
\caption{Single source domain generalization comparisons of PDE+ and baselines on PACS datasets based on ResNet-18~\cite{he2016deep}. The best result is highlighted in \textbf{boldface}.}
\scriptsize
\centering
\setlength{\tabcolsep}{2mm}
\begin{tabular}{lcccccc}
\toprule
Source & \multirow{2}{*}{Method} & \multicolumn{4}{c}{Target Domain} & \multirow{2}{*}{Avg} \\
\cmidrule(lr){3-6}
Domain & & Photo & Art & Cartoon & Sketch & \\
\midrule
\multirow{3}{*}{Photo} 
& ERM &-& 21.33 & 22.31 & 28.35 & 24.00 \\
& Augmix &-&26.90& 24.10&27.05&26.02\\
& PDE+ &-&\bf 25.43 & \bf 28.58& \bf 37.69& \bf 30.57\\
\midrule
\multirow{3}{*}{Art} 
& ERM & 47.54 &-& 34.51 & 34.48& 38.85\\
& Augmix & 51.37 &-& 42.06 & 36.75 & 43.40\\
& PDE+ & \bf 53.11 &-& \bf 43.90 & \bf 41.28 & \bf 46.10\\
\midrule
\multirow{3}{*}{Cartoon} 
& ERM & 43.59 & 29.78 &-& 33.87 & 35.75 \\
& Augmix & 45.74 & 30.81 &-& 37.31 &37.96 \\
& PDE+ & \bf 48.68 & \bf 33.00 &-& \bf 40.01 & \bf 40.57 \\
\midrule
\multirow{3}{*}{Sketch} 
& ERM & 18.74 & 16.16 & 25.26 &-&20.05\\
& Augmix & 26.28 & 26.51 & 45.34 &-&32.72\\
& PDE+ & \bf 30.05& \bf 30.90 & \bf 45.43&-&\bf 35.47\\
\bottomrule
\end{tabular}
\label{Tab:DG_PACS}
\end{wraptable}

\paragraph{Experiments Settings}
A brief introduction of datasets, baselines and metrics is provided here, details can be found in \cref{App:Settings}. 
\textbf{(1) Datasets}: Our experiments primarily focus on two types of datasets:
(\textit{\romannumeral 1}) The original and 15 shift corruption distributions provided by CIFAR-10(C), CIFAR-100(C) and Tiny-ImageNet(C)~\cite{krizhevsky2009learning,le2015tiny,hendrycks2018benchmarking}. 
(\textit{\romannumeral 2}) The PACS dataset \cite{li2017deeper} encompasses four different domains: photo, art, cartoon, and sketch. 
\textbf{(2) Baselines}: we consider the representative and SOTA methods as baselines: standard training; Lipschitz continuity based gradient regularization~\cite{165600}; Noise injection based methods, including EnResNet~\cite{doi:10.1137/19M1265302}, RSE~\cite{Liu_2018_ECCV}, NFM~\cite{lim2022noisy}; Data augmentation based methods, including Gaussian noise, Mixup~\cite{zhang2018mixup}, DeepAug~\cite{9710159}, AutoAug~\cite{Cubuk_2019_CVPR} and AugMix~\cite{hendrycks*2020augmix}; Adversarial training based methods, including PGD~\cite{madry2018towards} and RLAT~\cite{kireev2022on}. 
\textbf{(3) Metrics}: Accuracy is adopted as the main evaluation metric. Especially, for various corrupted distributions, mCE~\cite{hendrycks2018benchmarking} is adopted for two severity levels: severity across all levels and the severest level 5. 
More comprehensive results for other severity and metrics are in \cref{App:Experiments}.
\textbf{(4) Others}: According to \cref{Sec4:Adaptive_Distributional_Diffusion_Block}, the semantically similar samples in PDE+ are generated using AugMix~\cite{hendrycks*2020augmix}, a widely adopted data augmentation strategy that combines 7 distinct types of augmentations. It is important to note that we avoid 
overlap between these augmentations and the test distributions for most experiments.

\subsection{Q1: PDE+ Outperforms SOTA on Benchmarks}

\cref{Tab:cifar_ood} illustrate the results of PDE+ on CIFAR10(C), CIFAR100(C) and Tiny ImageNet(C) compared to baselines. ``*'' indicates that we reuse the results from~\cite{erichson2022noisymix,kireev2022on}.
``-'' indicates that this setting was not included in the paper. 
On original datasets, PDE+ achieves better performance than ERM, indicating that our diffusion does not obtain o.o.d. generalization at the cost of damaging performance on the original training distribution. 
The test distributions in corrupted datasets are different from training ones, which can be used to verify the effectiveness of generalization. Compared to numerous baselines across multiple categories, PDE+ achieves the best performance with respect to Acc and mCE on corruptions at the severest level and across all levels. The improvements are up to 3.8\% in Acc and 7.7\% in mCE. 
Such significant improvements make PDE+ stand out from other approaches that struggle to consistently improve performance across both original and diverse shifted distributions.

\begin{figure}[!t]
    \centering
    \small
    \includegraphics[width=1.0\linewidth]{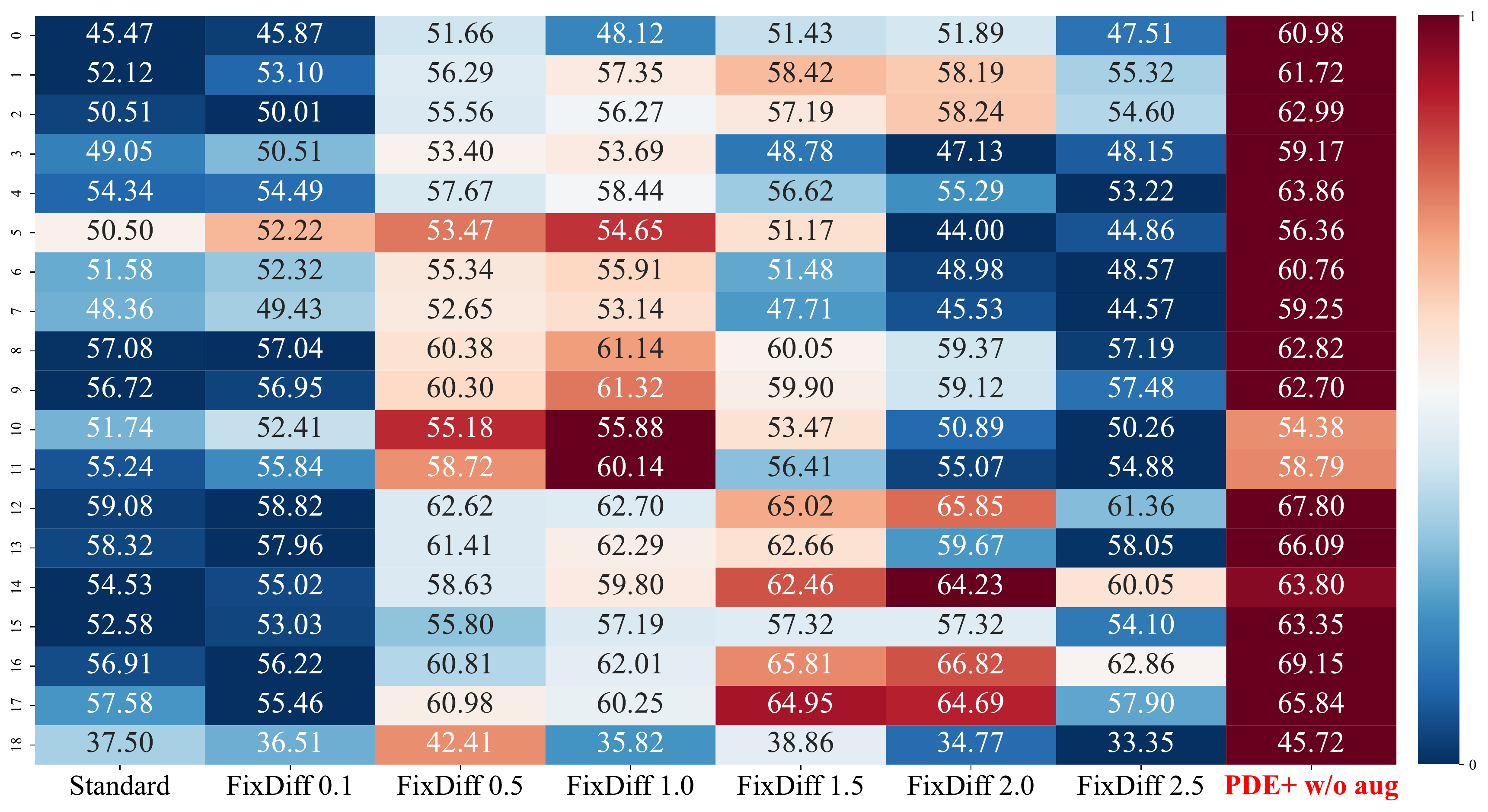}
    \caption{The heatmap of performance on neural networks with fixed diffusion (FixDiff) and PDE+ under fair comparison. The columns represent various training methods, with the FixDiff scale incrementally increasing from 0 to 2.5 across the first seven columns, while the last column is our PDE+. Each row corresponds to a unique test data distribution from the CIFAR-10-C dataset.}
    \label{Fig:Optimal}
\end{figure}

\cref{Tab:DG_PACS} illustrate results of PDE+ on PACS datasets. When training on a single source domain and testing on the remaining three domains, PDE+ surpasses baselines across all splits. This validates its efficacy not only in corrupted data, where distribution shifts may be relatively close, but also demonstrates effectiveness with cross-domain data where distribution shifts can be notably larger.

\begin{figure}[!t]
    \centering
    \small
    \includegraphics[width=1.0\linewidth]{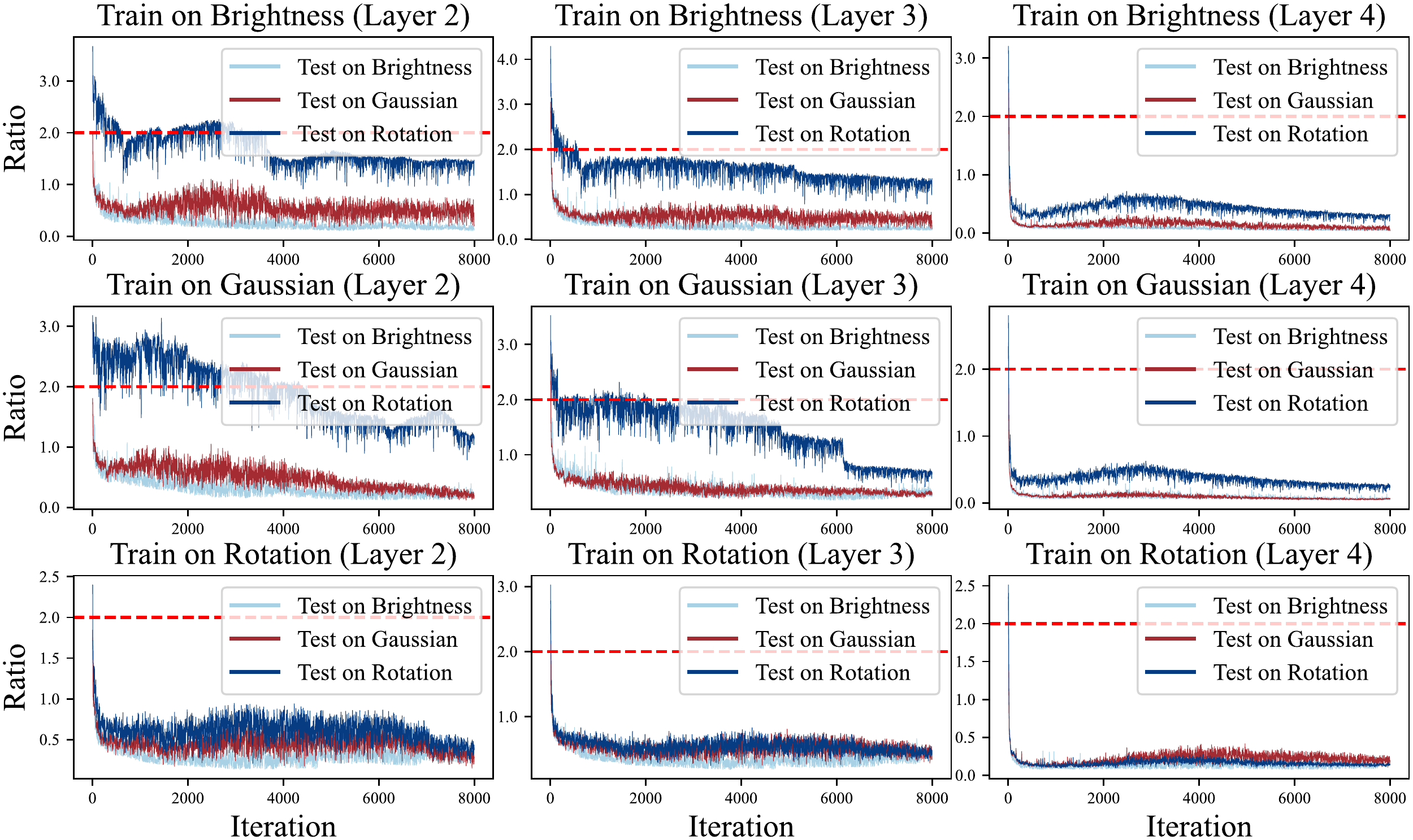}
    \caption{Diffusion coverage for unobserved distributions. Rows represent the training augmentations. Columns correspond to the layers of neural network. Each sub-figure includes three plots of distance-$\sigma$ ratio during training for test samples generated by different test augmentations.}
    \label{Fig:Ratio}
\end{figure}

\subsection{Q2: PDE+ Learns Appropriate Diffusion} \label{Exp:Diffusion_Coverage}

This experiment is devoted to evaluating whether our proposed approach, whose diffusion scale is guided by augmented samples, can learn the appropriate diffusion scope. 
For a fair comparison, we do not conduct diffusion for augmented samples and only use augmented samples for the diffusion coverage guidance of the original samples (PDE+ w/o aug). This experiment can be viewed as the ablation study to evaluate if our learnable diffusion really works compared to fixed-scale diffusion (FixDiff for shorthand). 
Two conclusions can be drawn from the experimental results shown in \cref{Fig:Optimal}: 
(1) Different corruption types, i.e., different distribution, prefers different magnitude/scale of smoothness, and a hard-to-please-everyone dilemma is caused by the fixed scale. 
(2) PDE+ indeed learns the appropriate diffusion scale. As is shown in the rightmost column, we can either achieve or be close to, the best performance of all corruption types.

\begin{figure}
    \centering
    \includegraphics[width=1.0\linewidth]{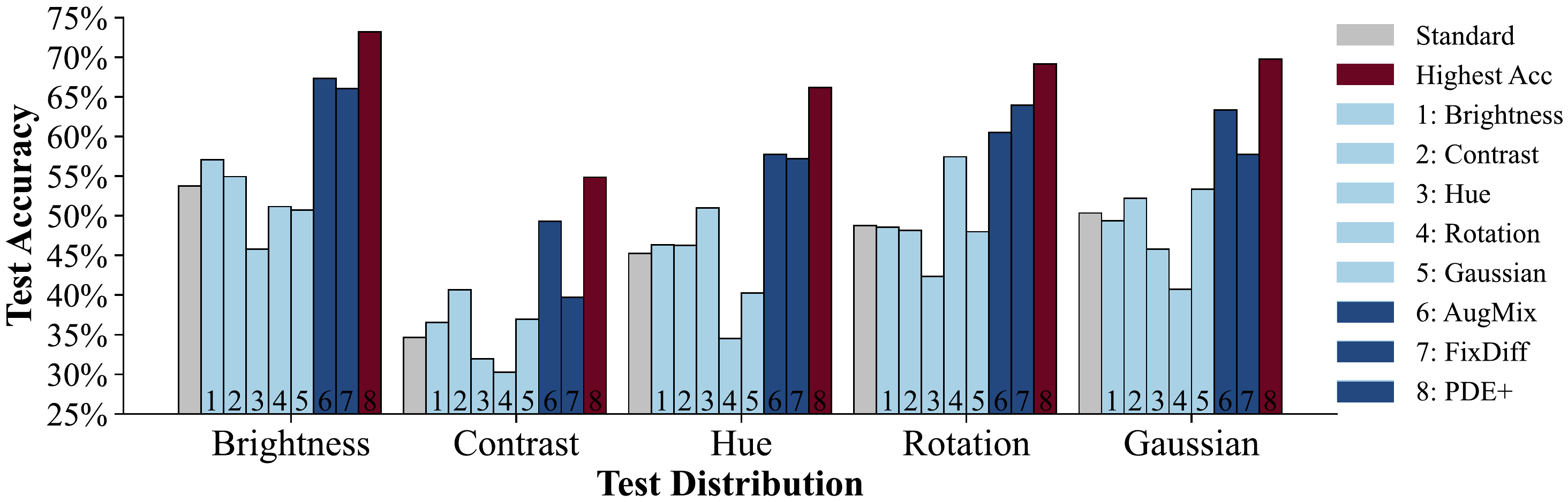}
    \caption{Generalization performance under five test distributions. The model is trained by different methods. The bar representing the highest performance is highlighted in red.}
    \label{Fig:Unseen}
\end{figure}

\subsection{Q3: PDE+ Generalizes Beyond Observation}

This subsection aims to demonstrate that our method can generalize on distributions beyond training ones.
\cref{Fig:Ratio} presents the changing trend of diffusion coverage for unobserved distributions, i.e., probabilities of unobserved test samples within the training diffusion distribution. This experiment is based on 2-$\sigma$ rule of Gaussian distribution, detailed description can be found in \cref{APP:Prob}. 
The results imply that even when training occurs on a single augmentation differing from testing, the likelihood of test samples being perceived as normal within the training diffusion distribution increases over time.
\cref{Fig:Unseen} represent the experiment as an extension of the previous one on \cref{Fig:motivation}. 
Notably, PDE+ outperforms all other augmentations, including AugMix, demonstrating its  capability of generalization on unobserved distributions.

\section{Conclusion}
In conclusion, we present a novel partial differential equations (PDE)-driven approach to address the generalization issue of neural networks across unseen data distributions, focusing on overcoming the limitations of data-driven methods. By modeling neural networks as solutions to PDEs in a transport equation framework, the connection between the solution smoothness of PDEs and the generalization of neural networks is established. The introduction of an adaptive distributional diffusion term helps improve the generalization of neural networks. An instantiation of this framework, called PDE+ can enhance the generalization via taking the augmented samples as semantic similar samples to guide the learning of adaptive distributional diffusion. Experimental results demonstrate the superior performance of PDE+ across various shifted distributions. This work opens up new avenues for research in generalization of neural networks from the PDE perspective and offers a promising direction for enhancing the generalization of neural networks.

\section*{Acknowledgments}
This work was supported by the National KeyR\&D Program of China (2022YFB3103700, 2022YFB3103704), the National Natural Science Foundation of China (NSFC) under Grants No.U21B2046 and No.62202448.

\bibliography{neurips_2023}
\bibliographystyle{abbrv}

\appendix
\section{Appendix Summary}
The appendix contains the following sections:
\begin{itemize}
    \item [(1)] Detailed related work (\cref{App:Related_Work}).
    \item [(2)] Proofs: 
    \begin{itemize}
        \item Proof for \cref{Them:TEGen} (\cref{App:Proof_Th1})
        \item Proof for \cref{Coro:TEGen} (\cref{App:Proof_Co1})
        \item Proof for \cref{Them:Feynman-Kac} (\cref{App:Proof_Th2})
    \end{itemize}
    \item [(3)] Algorithm (\cref{App:Algorithm}).
    \item [(4)] Additional Experiments and Analysis: 
    \begin{itemize}
        \item Comprehensive results for each type of corruption (\cref{APP:Detailed_Results_for_Every_Corruptions})
        \item Study on increasing severity (\cref{App:Detailed_Results_for_Increasing_Severity})
        \item Training curves, including diffusion coverage, loss, and accuracy (\cref{App:Training_Trends})
        \item Analysis for Diffusion Coverage of Unseen Distributions (\cref{APP:Prob})
    \end{itemize}
    \item [(5)] Settings: 
    \begin{itemize}
        \item The summary of datasets (\cref{App:Datasets})
        \item The baselines (\cref{App:Baselines})
        \item The metrics (\cref{App:Metrics})
        \item The hyper-parameters (\cref{App:Hyper_Parameters})
        \item The computing resources (\cref{App:Resources})
    \end{itemize}
    \item [(6)] The limitations and future explorations (\cref{App:Limitation}).
\end{itemize}

\section{Detailed Related Work}
\label{App:Related_Work}

\subsection{Data Augmentation}
Data augmentation is a widely adopted technique for enhancing the generalization performance of machine learning models. By applying various transformations to input data, data augmentation effectively increases the size of the training dataset, encouraging models to learn more generalized features.
A general form is shown in \cref{Eq:Data_Augmentation}, where $\theta$ and $L$ denote the parameter and loss, $\mathcal{D}$ denotes the training dataset, $(x,y)$ denotes the original data and its labels, $x^{\prime}$ denotes the augmented data.
Classical data augmentation techniques include flipping, rotation, cropping, and color jittering. 
More recent methods have been proposed to further improve generalization. 
Mixup~\cite{zhang2018mixup} interpolates between both feature and label  of two samples;
CutMix~\cite{Yun_2019_ICCV} combines segments of two images in a patch-wise manner;
DeepAug~\cite{9710159} employs reinforcement learning  to learn augmentation strategies from data; 
AugMix~\cite{hendrycks*2020augmix} mixes augmented images and enforces consistent embeddings of the augmented images, which serves as a strong data augmentation technique for generalization.
Although data augmentation is an intuitive yet powerful method, it is limited by its data-driven nature, as the choice of specific augmentation strategies remains a limitation in terms of generalizaiotn across distributions. 
For example, a Gaussian augmentation strategy may struggle when applied to images with illumination change. 
Even with the combination of multiple augmentation strategies, the generalization is still unassured for distribution outside the combination. 
\begin{equation}
\min _\theta \mathbb{E}_{(x, y) \sim \mathcal{D}}\left[ L(\theta, x^{\prime}, y)\right]
\label{Eq:Data_Augmentation}
\end{equation}

Our method PDE+, serving as a global smoothness constraint via the lens of PDE, 
can be directly understood as follows:
Based on the smoothness assumption, our mechanism smooths the gap between original sample and augmented sample with random sampled noise. In other words, we create a more continuous and smoothed surrounding from the original sample to cover the augmented sample. PDE+ enables better coverage of potentially unobserved distributions within the gap or surrounding, thereby improving generalization beyond data augmentation.

\subsection{Adversarial Training}

Adversarial Training (AT) is a robust optimization technique designed to improve adversarial robustness, i.e., the robustness of machine learning models against adversarial examples, which are small but maliciously perturbed inputs that can deceive models into making incorrect predictions. 
A general form is shown in \cref{Eq:Adversarial_Training}, where $\delta$ and $\mathcal{S}$ denote the perturbation and its radius.
The concept was initially introduced by ~\cite{DBLP:journals/corr/GoodfellowSS14}, and several AT approaches have been proposed over the years, including Fast Gradient Sign Method (FGSM)~\cite{DBLP:journals/corr/GoodfellowSS14}, Projected Gradient Descent (PGD)~\cite{madry2018towards}, and TRADES~\cite{pmlr-v97-zhang19p}. 
Although AT can significantly improve generalization on adversarial examples, two phenomena have been observed: (1) It exists an inherent trade-off between the generalization on adversarial examples and original clean samples, which has been widely observed in~\cite{tsipras2018robustness,pmlr-v97-zhang19p}. (2) AT enforces models robust against adversarial attacks of a specific type and certain magnitudes, its gained robustness does not extrapolate to larger perturbations nor unseen attack types~\cite{chen2021robust}. 
These observed phenomena underscore the limitations inherent in the data-driven nature of AT. In essence, the model  merely fit, or even overfit, the training examples it has observed~\cite{raghunathan2020understanding,chen2021robust}. This notion aligns with the proposition that the trade-off between standard and robust error is likely to diminish given an infinite dataset~\cite{raghunathan2020understanding}, or injecting more learned smoothness~\cite{chen2021robust}.
\begin{equation}
\min _\theta \mathbb{E}_{(x, y) \sim \mathcal{D}}\left[\max _{\delta \in \mathcal{S}} L(\theta, x+\delta, y)\right]
\label{Eq:Adversarial_Training}
\end{equation}

While the primary objective of AT is typically to enhance adversarial robustness, our aims diverge to concentrate on a broader concept of generalization across diverse distributions. There exist works that exploit AT to bolster generalization, and it has been discovered that the judicious selection of perturbation radius during AT can enhance non-adversarial generalization on common corruptions~\cite{kireev2022on}.
Our method, PDE+, can be construed as a synergy of "smoothness" and "appropriate radius selection", thus sharing common insights from these aforementioned works.

\subsection{Noise Injection}
Noise Injection is another technique aiming at improving the generalizatio of machine learning models. 
It involves injecting noise into input data~\cite{an1996effects}, activation functions~\cite{pmlr-v48-gulcehre16}, or hidden layers~\cite{NEURIPS2020_c16a5320}. Several noise types have also been explored, including Gaussian noise~\cite{doi:10.1137/19M1265302}, dropout noise~\cite{DBLP:journals/corr/abs-1812-00174}, and Bernoulli noise.
Recent works, such as ~\cite{kireev2022on,lim2022noisy}, have demonstrated that training models with noise can improve their generalization to unseen corruptions.
Existing methods are as follows, EnResNet~\cite{doi:10.1137/19M1265302} improves performance through an ensemble of ResNets with injected noise. RSE~\cite{Liu_2018_ECCV} presents a framework that injects noise and employs self-ensembling during testing to enhance model robustness. NFM~\cite{lim2022noisy} combines noise injection and manifold mixup~\cite{verma2019manifold} to expand the generalization capacity of the model. However, the magnitude of noise plays a significant role in determining the effectiveness of noise injection. It has been shown in ~\cite{kireev2022on} that noise injection tends to overfit to a particular magnitude of noise used for training, resulting in a significant detrimental effect on generalization, which is identified as $\sigma$-overfitting.
PDE+ conducts adaptive diffusion grounded in the principles of PDEs to ensure smoothness, avoiding overfitting to a particular magnitude.

\subsection{Lipschitz Continuity}
Lipschitz continuity (shown in \cref{Eq:Lipschitz}) is a mathematical property that serves as a popular constraint on the smoothness of functions, ensuring that the output of a function does not change too drastically with respect to small changes in its input~\cite{HagerLipSIAM}. Given two metric spaces $\left(X, d_X\right)$ and $\left(Y, d_Y\right)$, where $d_X$ denotes the metric on the set $X$ and $d_Y$ is the metric on set $Y$, a function $f: X \rightarrow Y$ is called Lipschitz continuous if there exists a real constant $K \geq 0$ such that, for all $x_1$ and $x_2$ in $X$,
\begin{equation}
d_Y\left(f\left(x_1\right), f\left(x_2\right)\right) \leq K d_X\left(x_1, x_2\right) \label{Eq:Lipschitz}
\end{equation}

In the field of deep learning, Lipschitz continuity has become a key aspect in improving model generalization.
Methods grounded in Lipschitz continuity can be broadly categorized into two groups: architectural constraints and regularization approaches. 
Architectural constraints place limitations on the operator norm, e.g., weight clipping~\cite{arjovsky2017wasserstein} controls the Lipschitz constant by setting bounds on the weights. Spectral normalization~\cite{miyato2018spectral} normalizes the spectral norm of the model's weight matrices.
These methods are proved to satisfy Lipschitz constraints. However, they may limit the expressive capacity of the model. 
For instance, it's been shown that norm-constrained ReLU networks cannot approximate simple functions such as the absolute value~\cite{huster2019limitations}. 
Regularization approaches impose smoothness by regularization. 
For instance, \cite{165600} imposes smoothness by applying regularization on gradients.
\cite{gulrajani2017improved} conducts gradient penalty to fix the poor gradient behaviors of weight clipping.
Although these methods show promising results in practice, they do not guarantee Lipschitz constraint on a global scale~\cite{anil2019sorting}.

\section{Proof} \label{App:Proof}

\subsection{Proof for \cref{Them:TEGen}} \label{App:Proof_Th1}

Recalling that in \cref{Them:TEGen} we have, Given TE with diffusion term (\cref{Eq:TE_FixDiff}) with terminal condition $u(\mathbf{x}, 1) = o(\mathbf{x})$, where $F(\mathbf{x}, \boldsymbol{\theta}(t))$ be a Lipschitz function in both $\mathbf{x}$ and $t$, $o(\mathbf{x})$ be a bounded function. Then, for any small $\delta$,
$|u(\mathbf{x}+\delta, 0)-u(\mathbf{x}, 0)| \leq C\big(\frac{\|\delta\|_2}{\sigma}\big)^\alpha$ holds for constant $\alpha>0$ if $\sigma \leq 1$, where
$\|\delta\|_2$ is the $\ell_2$ norm of $\delta$, and $C$ is a constant that depends on $d,\|o\|_{\infty}$, and $\|F\|_{L_{\mathbf{x}, t}^{\infty}}$.

\begin{proof}
For simplicity, here we only illustrate the case for $d=1$ in the initial value problem (i.e., $t \in (0,T)$ given $u(\mathbf{x},0)$), where the terminal value problem can be proved by reverse time. General proof and the details of following notations could be found in \cite{ladyzhenskaya1968linear}.
The outline of this proof is as follows: First, we adopt the Sobolev embedding theorem for embedding $H^k$ space into Hölder spaces, which constructs the left-hand side of the main inequality in \cref{Them:TEGen}. Consequently, the remainder of the proof applies energy method to give an $H^k$ estimate of solution $u$.

According to Sobolev embedding theorem~\cite{brezis2011functional}, for $\alpha < k-\frac{d}{2}$, it holds that
\begin{equation}
\label{eq:sobolev}
    \sup_{\mathbf{x},\delta} \frac{\| u(\mathbf{x} + \delta,T) - u(\mathbf{x},T) \|}{\| \delta \|_2^{\alpha}} \leq C_1 \| u(\mathbf{x},T) \|_{H^{k}}
\end{equation}
where constant $C_1$ depends on $d$ and $k$. For $d=1$, it suffices to prove \cref{eq:sobolev} for $k=1$. Multiplying \cref{Eq:TE_FixDiff} by $u$ and integrating it for variable $\mathbf{x}$, we obtain
\begin{equation}
\begin{split}
    & \frac{\mathrm{d}}{\mathrm{d}t} \| u \|_{L^2}^2 + 2 \sigma \| \nabla u \|_{L^2}^2  \leq 2 \| F \|_{L^{\infty}_{\mathbf{x},t}} \| \nabla u \|_{L^2} \| u \|_{L^2} \\
    & \leq \sigma \| \nabla u \|_{L^2}^2 + \frac{\| F \|_{L^{\infty}_{\mathbf{x},t}}^2 \| u \|_{L^2}^2}{\sigma} \\
\end{split}
\end{equation}
which implies
\begin{equation}
\label{eq:Th1energy}
    \frac{\mathrm{d}}{\mathrm{d}t} \| u \|_{L^2}^2 + \sigma \| \nabla u \|_{L^2}^2 \leq \frac{\| F \|_{L^{\infty}_{\mathbf{x},t}}^2 \| u \|_{L^2}^2}{\sigma}
\end{equation}
Then by Grönwall's inequality, $\| u \|_{L^2}$ satisfies $\| u(\mathrm{x},t)\|_{L^2}^2 \leq e^{\| F \|_{L^{\infty}_{\mathbf{x},t}}^2 t / \sigma } \| u(\mathbf{x},0)\|_{L^2}$. Plugging the estimate of $\| u \|_{L^2}$ back to \cref{eq:Th1energy}, it is easy to see
\begin{equation}
    \| \nabla u (\mathbf{x},t) \|_{L^2}^2 \leq \frac{\| F \|_{L^{\infty}_{\mathbf{x},t}}^2}{\sigma^2} e^{\| F \|_{L^{\infty}_{\mathbf{x},t}}^2 t / \sigma } \| u(\mathbf{x},0)\|_{L^2}
\end{equation}
which suggests
\begin{equation}
\label{eq:uh1norm}
    \| u (\mathbf{x},t) \|_{H^1}^2 \leq \frac{\| F \|_{L^{\infty}_{\mathbf{x},t}}^2+1}{\sigma^2} e^{\| F \|_{L^{\infty}_{\mathbf{x},t}}^2 t / \sigma } \| u(\mathbf{x},0)\|_{L^2}
\end{equation}
Plugging \cref{eq:uh1norm} into \cref{eq:sobolev} yields
the desired statement.

\end{proof}

\subsection{Proof for \cref{Coro:TEGen}}
\label{App:Proof_Co1}
Recalling that in \cref{Coro:TEGen} we have, Generalization Error (GE) of model $u(\mathbf{x},0)$ trained on training set $s_N$ is upper bounded by diffusion $\sigma$. For any $\epsilon > 0$, the following inequality holds with probability at least $1-\epsilon$.

\begin{equation*}
\footnotesize \mathrm{GE}\left(u(\mathbf{x},0), s_N \right) \leq C \cdot L \left(\frac{\|\delta'\|_2}{\sigma}\right)^\alpha \! +M \sqrt{ \frac{2K\ln 2+2 \ln (1 / \epsilon)}{N}}
\end{equation*}

\begin{proof}
Assume that the loss function is denoted as $l(y, f(x))$, which quantifies the discrepancy between the true label $y$ and the predicted label by the model $f(x)$. A training set is denoted as $s_N$, where $\{(x_{s_i}, y_{s_i})\}_{i=1}^N \in s_N$ are $N$ samples drawn from the entire data distribution $\mathcal{D}$.

Let $\mathcal{A}_{s_N}$ be a learning algorithm trained on the training set $s_N$. The empirical risk of $\mathcal{A}_{s_N}$ (\cref{Eq:REmp}), while the expected risk of $\mathcal{A}_{s_N}$ on the whole data distribution $\mu$ is defined in \cref{Eq:RExp}. Consequently, the definition for generalization (generalization error~\cite{DBLP:conf/colt/XuM10}), which measures the difference between empirical risk and expected risk, is shown in \cref{Eq:GE}.
\begin{align}
\ell_{emp} (f, s_N ) & \triangleq \frac{1}{N} \sum^N_{i=1} \ell (y_{s_i}, f (x_{s_i})) \label{Eq:REmp}\\
\ell_{exp} (f) & \triangleq \mathbb{E}_{(x, y) \sim \mu}[\ell(y, f(x))] \label{Eq:RExp}\\
GE(f, s_N) & \triangleq | \ell_{exp}(f)- \ell_{emp} (f, s_N )| \label{Eq:GE}
\end{align}
In our case, $f(x)$ is equivalent to the solution of TE, i.e., $u(x,0)$. Therefore, our generalization error is defined as:
\begin{align*}
 GE\left(u(x,0), s_N \right) \triangleq \left|\ell_{exp}\left(u(x,0),y\right)-\ell_{emp}\left(u(x_s,0),y_{s} \right)\right| \label{Eq:GE_u}
\end{align*}
Referring back to \cref{Them:TEGen}, similar to \cite{DBLP:conf/colt/XuM10}, for a $K$-way classification problem, we partition the data space $\mathcal{Z}$ into $K$ partitions $\{P_i\}_{i=1}^K$, where $P_i$ signifies the partition for the $i$th class. Let $\delta'$ denote the maximum potential perturbation for each partition, and $l$ represent the $L$-Lipschitz loss function which is upper bounded by $M$. Therefore, for any $x_s$ and $x$ in the same class $P_i$, \cref{Eq:smoothness} holds.
\begin{align}
&|\ell(u(x, 0),y)-\ell(u(x_s, 0),y_s)| \\
&\leq C\cdot L\left(\frac{\|x-x_s\|_2}{\sigma}\right)^\alpha 
\leq C\cdot L\left(\frac{\|\delta'\|_2}{\sigma}\right)^\alpha
\label{Eq:smoothness}
\end{align}
We can derive a diffusion coefficient upper bound to generalization error. Let $N_i$ be the set of training samples of $s_N$ that fall into the $P_i$. Due to Breteganolle-Huber-Carol inequality\cite{DBLP:conf/colt/XuM10}, \cref{Eq:BHC} holds with probability at least $1-\epsilon$.
\begin{align}
\sum_{i=1}^K\left|\frac{\left|N_i\right|}{n}-\mu\left(C_i\right)\right| \leq \sqrt{\frac{2 K \ln 2+2 \ln (1 / \epsilon)}{N}} \label{Eq:BHC}
\end{align}
The derivation is as follows, where (a), (b), and (c) are due to the triangle inequality, the definition of $N_i$, and the smoothness bound from \cref{Them:TEGen} of the original paper. (d) is due to \cref{Eq:BHC}, which connects generalization with diffusion.
\begin{align*}
\tiny
&GE\left(u(x,0), s_N \right) \triangleq \left|\ell_{exp}\left(u(x,0),y\right)-\ell_{emp}\left(u(x_s,0),y_{s} \right)\right|\\
&=\left|\sum_{i=1}^K \mathbb{E}\left(\ell\left(u(x,0), y\right) \mid (x,y) \in P_i\right) \mu\left(P_i\right)
-\frac{1}{N} \sum_{i=1}^N \ell\left(u(x_{s_i},0), y_{s_i}\right)\right| \\
& \stackrel{(a)}{\leq}\left|\sum_{i=1}^K \mathbb{E}\left(l\left(u(x,0), y\right) \mid (x,y) \in P_i\right) \frac{\left|N_i\right|}{N}-\frac{1}{N} \sum_{i=1}^N l\left(u(x_{s_i},0), y_{s_i}\right)\right| \\
&+\left|\sum_{i=1}^K \mathbb{E}\left(l\left(u(x,0), y\right) \mid (x,y) \in P_i\right) \mu\left(P_i\right)-\sum_{i=1}^K \mathbb{E}\left(l\left(u(x,0), y\right) \mid (x,y) \in P_i\right) \frac{\left|N_i\right|}{N}\right| \\
& \stackrel{(b)}{\leq}\left|\frac{1}{N} \sum_{i=1}^K \sum_{j \in N_i} \max_{(x,y) \in P_i} \mid l\left(u(x_{s_j},0), y_{s_j}\right)-l\left(u(x,0), y\right) \mid\right|+\left| \max_{(x,y) \in \mathcal{Z}} \mid l\left(u(x,0), y\right)\mid \sum_{i=1}^K \mid \frac{\left|N_i\right|}{N}-\mu\left(P_i\right)\mid\right| \\
&\stackrel{(c)}{\leq} C \cdot L \left(\frac{\|\delta'\|_2}{\sigma}\right)^\alpha + M \sum_{i=1}^K\left|\frac{\left|N_i\right|}{N}-\mu\left(P_i\right)\right|
\stackrel{(d)}{\leq} C \cdot L  \left(\frac{\|\delta'\|_2}{\sigma}\right)^\alpha +M \sqrt{ \frac{2K\ln 2+2 \ln (1 / \delta)}{N}}
\end{align*}
\end{proof}

\subsection{Proof for \cref{Them:Feynman-Kac}}
\label{App:Proof_Th2}

Recalling that in \cref{Them:Feynman-Kac} we have, TE with adaptive distributional diffusion term (\cref{Eq:TE_AdaDiff})can be solved using the Feynman-Kac formula~\cite{10.2307/1990512}, The result is shown in \cref{Eq:FK_U,Eq:FK_SDE}, where $B_t$ represents the Brownian motion~\cite{PhysRev.36.823}. 
\begin{align}
u(\hat{\mathbf{x}}, 0) &= \mathbb{E}\left[o(\mathbf{x}(1)) \mid \mathbf{x}(0)=\hat{\mathbf{x}}\right]\\
\mathrm{d} \mathbf{x}(t) &= F(\mathbf{x}(t), \boldsymbol{\theta}(t))\, \mathrm{d} t+ G(\mathbf{x}(t), \boldsymbol{\phi}(t)) \cdot \mathrm{d} B_t
\end{align}

\begin{proof}
Assume that
\begin{align}
\label{eq:dxtFG}&\mathrm{d}\mathbf{x}_t = F(\mathbf{x}_t,t)\mathrm{d}t + G(\mathbf{x}_t,t)\mathrm{d} B_t \\
\label{eq:xrangetT}&\mathbf{x}_t = \mathbf{x}, t < T
\end{align}

Let $u(\mathbf{x},t) = \mathbb{E}[o(\mathbf{x}_T)|\mathcal{F}_t]$. \cref{eq:dxtFG} shows $\mathbf{x}_t$ is a Markov process, and therefore $u(\mathbf{x},t) = \mathbb{E}[o(\mathbf{x}_T)|\mathbf{x}_t=\mathbf{x}]$. Moreover, 
\begin{equation}
\footnotesize
\begin{split}\label{eq:longderivation}
    u(\mathbf{x},t)& =\mathbb{E}[o(\mathbf{x}_T)|\mathcal{F}_t]  \\
    &=\mathbb{E}[\mathbb{E}[o(\mathbf{x}_T)|\mathcal{F}_{t+\mathrm{d}t}]|\mathcal{F}_t] \\
    &=\mathbb{E}[u(\mathbf{x}_{t+\mathrm{d}t},t+\mathrm{d}t)|\mathcal{F}_t] \\
    &=\mathbb{E}[u(\mathbf{x}_t,t)+\frac{\partial u}{\partial t}\mathrm{d}t + \frac{\partial u}{\partial \mathbf{x}}\mathrm{d} \mathbf{x}_t + \frac{1}{2}\frac{\partial^2 u}{\partial\mathbf{x}^2}\mathrm{d}[\mathbf{x},\mathbf{x}](t)|\mathcal{F}_t] \\
    &=u(\mathbf{x},t)+\frac{\partial u}{\partial t}\mathrm{d}t + \frac{\partial u}{\partial \mathbf{x}}\mathbb{E}[\mathrm{d}\mathbf{x}_t|\mathcal{F}_t] + \frac{1}{2}\frac{\partial^2 u}{\partial \mathbf{x}^2}\mathbb{E}[\mathrm{d}[\mathbf{x},\mathbf{x}](t)|\mathcal{F}_t]
\end{split}
\end{equation}

According to \cref{eq:dxtFG}, the above equation yields
\begin{align}
    \frac{\partial u}{\partial t} + F(\mathbf{x},t)\frac{\partial u}{\partial\mathbf{x}} + \frac{1}{2}G^2(\mathbf{x},t)\frac{\partial^2 u}{\partial \mathbf{x}^2} = 0
\end{align}
\end{proof}

\section{Algorithm} \label{App:Algorithm}
The following pseudocodes provide an overview of PDE+. The training process is described in \cref{Alg:Train}, the testing process is described in \cref{Alg:Test}.

\begin{algorithm}
\caption{Training Phase of PDE+}
  \label{Alg:Train}
  \KwIn{Training dataset $(\mathbf{x}, y)$. Number of blocks $L$. Data Augmentor $\mathrm{Augmentor}$}
  \KwOut{Trained parameters $\boldsymbol{\theta}$, $\boldsymbol{\phi}$ and $\boldsymbol{\psi}$}

  \While{epoch $\leq$ MAX\_ITER}{

    \tcc{forwarding for every layer's optimal scale}
  
    $\mathbf{x}^a=\mathrm{Augmentor}(\mathbf{x})$
       
    $\tilde{\mathbf{h}}_0 = \mathbf{x}$ 
    
    $\mathbf{h}^a_0 = \mathbf{x}^a$

    \For{$l=1,2, \dots, L$ }{
        
        \tcc{do convection for original data and augmented data}

        $\mathbf{h}_l=f_{\boldsymbol{\theta}}(\mathbf{\tilde{h}}_{l-1})$
    
        $\mathbf{h}^a_l=f_{\boldsymbol{\theta}}(\mathbf{h}^a_{l-1})$

        \tcc{do diffusion only for original data}

        $\boldsymbol{\sigma}_l=g_{\boldsymbol{\phi}_l}\left(\mathbf{h}_l\right)$
        
        $\mathbf{\tilde{h}}_l=\mathbf{h}_l+\boldsymbol{\sigma}_l \cdot \mathcal{N}(\mathbf{0}, \mathbf{I})$

        compute loss $- \log p(\mathbf{h}_l^a \mid \mathbf{h}_l)$ for every layers according to \cref{Eq:Obj_nll} to update $\boldsymbol{\phi}$.
             
    }
  
    \tcc{forwarding for primary task}
    $\mathbf{x} = \mathrm{concat}(\mathbf{x},\mathbf{x}^a)$ \label{Alg:Concat}

    $\mathbf{\tilde{h}}_L = u_{\boldsymbol{\theta},\boldsymbol{\phi}}\left(\mathbf{x}\right)$

    compute loss $- \log p(y \mid \mathbf{x})$ for the last layer according to \cref{Eq:Obj_cla} to update $\boldsymbol{\theta}, \boldsymbol{\phi}$ and  $\boldsymbol{\psi}$.   
  }
\end{algorithm}

\begin{algorithm}
  \SetAlgoLined
  \LinesNumbered %
  \KwIn{Testing dataset $\mathbf{x}$. Ensemble iter $E$}
  \KwOut{Prediction $\hat{y}$}
  \caption{Testing Phase of PDE+}
  \label{Alg:Test}

    initialize $\mathbf{h} = (0, 0, . . . , 0)$
    
    \tcc{do ensemble for multiple times}
    
    \For{$i=1,2 \dots E$ }{
        $\mathbf{\tilde{h}}_L = u_{\boldsymbol{\theta},\boldsymbol{\phi}}\left(\mathbf{x}\right)$

        $\mathbf{h} = \mathbf{h} + \mathbf{\tilde{h}}_L$
    }

    \tcc{predict label according to the max ensembling probability}
    $\hat{y}=\arg \max o_{\boldsymbol{\psi}}\left(\mathbf{h}\right)$
    
\end{algorithm}

\begin{sidewaystable*}[p]
\renewcommand{\arraystretch}{1.02}
\caption{
Comprehensive comparison of PDE+ and various baseline models on CIFAR-10, CIFAR-10-C and CIFAR-100, CIFAR-100-C. All evaluated models employ the ResNet-18 architecture as the foundation. 
Evaluations are based on Accuracy (\%) for each individual corruption, as well as Average Accuracy (Acc \%), Mean Corruption Error (mCE \%), and Relative Mean Corruption Error (rmCE \%) for overall performance. 
The reported performance of our PDE+ reflects the average across five runs with varying seeds, with a maximum standard deviation under 0.1\%.
The most notable results are indicated with \textbf{boldface} for the top performance, and \ul{underline} for the second-best results achieved by our PDE+.}
\centering
\begin{small}
\setlength{\tabcolsep}{1.5mm}{
\begin{tabular}{clccccccccccccccc|ccc}
\toprule
& \multirow{2}*{Method} & \multicolumn{3}{c}{Whether} & \multicolumn{4}{c}{Blur} & \multicolumn{4}{c}{Noise} & \multicolumn{4}{c}{Digital} & \multicolumn{3}{c}{Avg} \\
\cmidrule(lr){3-5} \cmidrule(lr){6-9} \cmidrule(lr){10-13} \cmidrule(lr){14-17} \cmidrule(lr){18-20}
& & Snow & Fog & Frost & Glass & Defocus & Motion & Zoom & Gaussian & Shot & Impulse & Pixel & Bright & Contrast & JPEG & Elastic & Acc($\uparrow$) & mCE($\downarrow$) & rmCE($\downarrow$) \\
\midrule
\multirow{9}{*}{\rotatebox{90}{CIFAR-10(-C)}} 
& ERM & 83.20 & 89.35 & 80.05 & 55.02 & 82.74 & 78.48 & 78.79 & 48.15 & 60.63 & 53.04 & 75.68 & 94.04 & 76.34 & 78.97 & 85.05 & 74.63 & 100 & 100 \\
& GradReg & 82.69 & 84.50 & 82.65 & 59.61 & 83.38 & 78.86 & 80.19 & 63.80 & 72.27 & 64.88 & 79.97 & 92.14 & 70.18 & 84.81 & 84.51 & 77.62 & 96.29 & 86.09 \\
& AutoAug & 87.45 & 92.56 & 88.23 & \textbf{80.32} & 91.15 & 85.41 & 89.21 & 66.91 & 75.45 & 81.82 & 81.70 & \textbf{95.16} & \textbf{94.58} & 83.76 & 86.95 & 85.37 & 61.74 & 50.48\\
& AugMix & 87.53 & 91.20 & 87.86 & 71.47 & 92.67 & 89.32 & 90.76 & 79.14 & 85.03 & 82.11 & 84.50 & 94.18 & 82.83 & 89.06 & 86.25 & 86.25 & 60.44 & 49.01\\
& PGD$_{\ell_{\infty}}$ & 86.68 & 75.11 & 86.73 & 76.33 & 86.51 & 81.15 & 85.13 & 81.89 & 85.18 & 72.23 & 89.73 & 91.63 & 57.60 & 90.18 & 86.58 & 82.17 & 86.53 & 77.90 \\
& PGD$_{\ell_2}$ & 86.68 & 77.45 & 86.85 & 76.99 & 87.12 & 82.53 & 86.18 & 81.28 & 84.90 & 72.34 & 89.87 & 92.27 & 63.86 & 90.38 & 87.46 & 83.08 & 81.06 & 72.07 \\
& RLAT & 86.13 & 76.62 & 87.15 & 79.94 & 82.38 & 86.02 & 84.69 & 84.69 & 87.10 & 76.42 & 89.52 & 91.84 & 62.50 & 90.39 & 87.39 & 83.67 & 80.98 & 66.71\\
& RLAT$_{\mathrm{AM}}$ & 88.28 & 88.37 & 89.18 & 79.40 & 92.67 & 90.03 & 92.03 & \textbf{85.97} & \textbf{89.22} & 84.39 & \textbf{90.45} & 93.66 & 79.79 & \textbf{90.47} & 90.32 & 88.28 & 55.60 & 40.20 \\
& PDE+ & \textbf{90.09} & \textbf{93.56} & \textbf{89.82} & 76.32 & \textbf{94.12} & \textbf{91.91} & \textbf{92.81} & \ul{83.59} & \ul{87.63} & \textbf{85.10} & 86.05 & \ul{94.84} & \ul{93.22} & 87.15 & \textbf{90.41} & \textbf{89.11} & \textbf{48.07} & \textbf{33.83}\\
\midrule
\multirow{9}{*}{\rotatebox{90}{CIFAR-100(-C)}} 
& ERM & 55.40 & 64.65 & 50.49 & 24.18 & 60.03 & 55.45 & 53.60 & 23.02 & 31.44 & 25.48 & 52.45 & 73.72 & 55.38 & 52.31 & 61.50 & 49.27 & 100 & 100 \\
& GradReg & 55.87 & 56.40 & 55.87 & 28.63 & 60.15 & 55.06 & 55.98 & 37.12 & 45.42 & 35.20 & 59.40 & 69.91 & 45.98 & 60.44 & 52.16 & 52.16  & 96.95 & 80.79\\
& AutoAug & 60.44 & 67.63 & 59.55 & 46.71 & 67.04 & 57.56 & 62.84 & 31.86 & 41.16 & 58.85 & 59.30 & 75.24 & \textbf{74.59} & 55.60 & 62.47 & 58.72 & 83.12 & 60.58\\
& AugMix & 62.70 & 67.19 & 59.73 & 42.94 & 76.14 & 67.79 & 70.54 & 48.38 & 55.85 & 57.50 & 62.86 & 73.60 & 60.30 & 61.92 & 67.42 & 61.93 & 77.51 & 56.50  \\
& PGD$_{\ell_{\infty}}$ & 59.61 & 47.17 & 59.55 & 45.86 & 61.52 & 55.63 & 59.55 & 48.42 & 53.83 & 35.37 & 67.05 & 67.84 & 36.37 & 66.59 & 61.15 & 55.03 & 93.49 & 68.73\\
& PGD$_{\ell_2}$ & 59.51 & 48.27 & 59.28 & 45.51 & 61.64 & 56.02 & 59.26 & 51.03 & 56.52 & 41.68 & 67.33 & 68.51 & 38.54 & \textbf{66.94} & 61.41 & 56.09 & 91.65 & 68.58\\
& RLAT & 57.81 & 47.72 & 57.73 & \textbf{50.73} & 60.79 & 55.01 & 58.33 & 50.73 & 59.62 & 46.97 & 66.39 & 66.42 & 37.65 & 66.08 & 60.92 & 56.54 & 91.98 & 64.16\\
& RLAT$_{\mathrm{AM}}$ & 62.45 & 60.36 & 50.49 & 48.61 & 70.98 & 66.94 & 69.19 & \textbf{56.68} & \textbf{61.76} & 56.08 & \textbf{69.40} & 71.55 & 52.94 & 61.98 & 66.70 & 62.77 & 77.38 & 49.66\\
& PDE+ & \textbf{66.68} & \textbf{69.84} & \textbf{63.15} & 46.99 & \textbf{76.50} & \textbf{73.08} & \textbf{74.38} & 50.80 & 58.81 & \textbf{61.48} & \ul{68.26} & \textbf{76.19} & \ul{66.57} & 61.59 & \textbf{69.98} & \textbf{65.62}  & \textbf{69.68} & \textbf{47.45} \\
\bottomrule
\end{tabular}}
\end{small}
\label{Tab:Full}
\end{sidewaystable*}

\section{Experiments and Analysis} \label{App:Experiments}
This section delves into an expanded set of experiments and associated analysis. These include (1) Comprehensive results for each type of corruption, as detailed in~\cref{APP:Detailed_Results_for_Every_Corruptions}; (2) Exploration of increasing severity impact, which can be found in~\cref{App:Detailed_Results_for_Increasing_Severity}; and (3) An examination of training trends, specifically those concerning diffusion coverage, loss, and accuracy, which are elucidated in~\cref{App:Training_Trends}.

\subsection{Comprehensive Results for Each Corruption Type}
\label{APP:Detailed_Results_for_Every_Corruptions}
This subsection presents the detailed results obtained by PDE+ and a range of baselines when tested on CIFAR-10-C and CIFAR-100-C. The evaluation metric includes accuracy for each individual corruption (all 15 types of corruptions), along with average accuracy, mCE and rmCE for all corruptions, as shown in \cref{Tab:Full}. 
From these results, we derive two key conclusions: (1) our proposed PDE+ exhibits superior performance across the majority of corruption types, and (2) our PDE+ improves performance in terms of all metrics, including accuracy, mCE and rmCE.

\subsection{Exploration of Increasing Severity Impact}
\label{App:Detailed_Results_for_Increasing_Severity}
\begin{figure}[!t]

    \centering
    \includegraphics[width=1.0\linewidth]{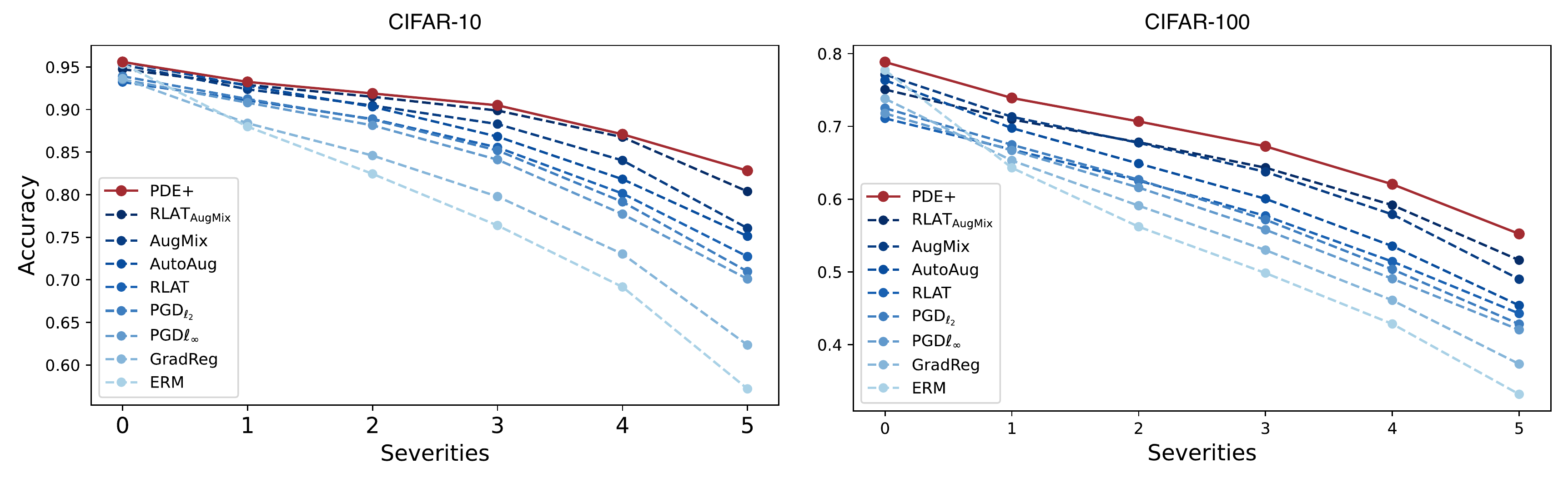}
    \caption{Performance comparison of PDE+ and various baselines on CIFAR-10, CIFAR-10-C, CIFAR-100, and CIFAR-100-C under increasing severity levels. All evaluated models employ the ResNet-18 architecture as the foundation. The x-axis denotes severity levels, with 0 symbolizing the uncorrupted (clean) versions of CIFAR-10 and CIFAR-100 datasets. The y-axis illustrates the accuracy of each model.}
    \label{Fig:Severity}

\end{figure}

This subsection provides analysis of the performance of PDE+ and various baselines on CIFAR10, CIFAR-10-C, CIFAR-100, and CIFAR-100-C datasets with increasing severity levels. The results are visually depicted in \cref{Fig:Severity}. From these results, we infer two main conclusions: (1) Our proposed PDE+ model consistently outperforms other methods across all severity levels, and (2) With an increase in the severity level, PDE+ demonstrates resilience and stability, unlike most baselines, which show a significant decline in performance. This effect is particularly evident in the case of the CIFAR-10 dataset.

\subsection{Training Curves}
\label{App:Training_Trends}
This subsection visualizes the training curves for PDE+ on CIFAR-10 and CIFAR-100, facilitating a comprehensive understanding of model behavior throughout the learning phase.

\begin{figure}[!t]

    \centering
    \includegraphics[width=1.0\linewidth]{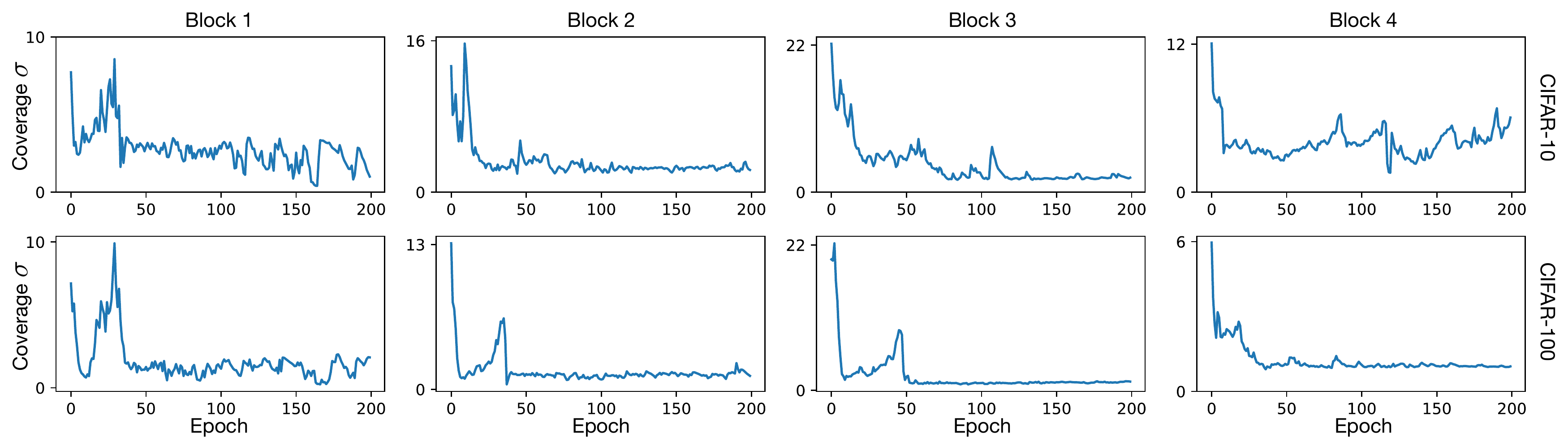}
    \caption{Evolution of diffusion coverage for PDE+ on CIFAR-10 and CIFAR-100 throughout the training phase. The employed backbone is ResNet-18. The x-axis denotes the progression in training epochs, while the y-axis depicts the mean value of diffusion coverage, symbolized as $\sigma$.}
    \label{Fig:Scale}

\end{figure}
\begin{figure}[!t]
    \centering
    \includegraphics[width=1.0\linewidth]{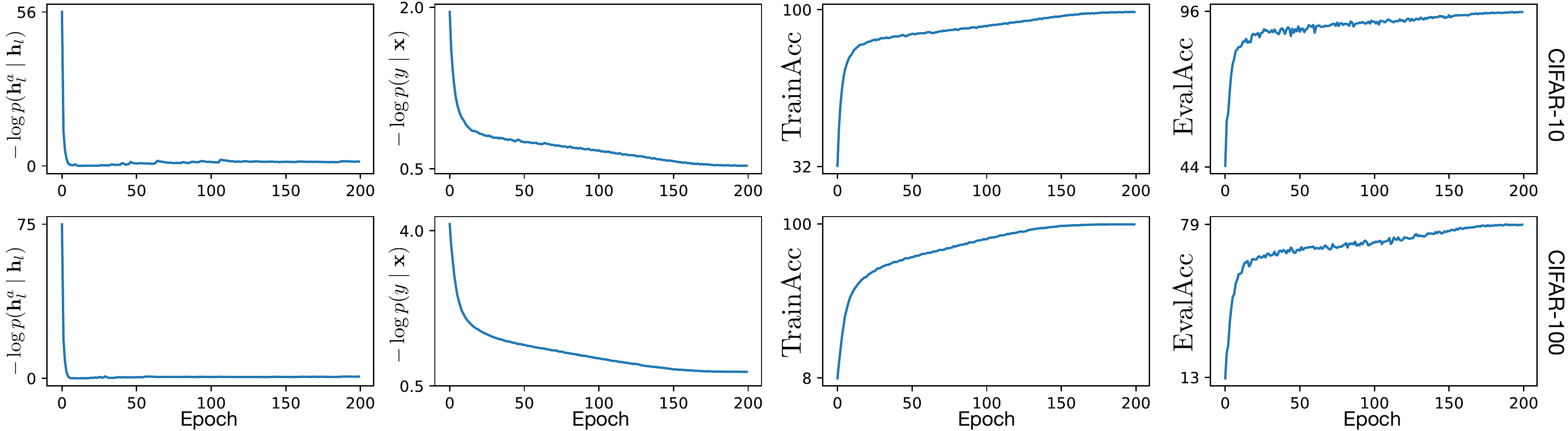}
    \caption{Evolution of loss and accuracy for PDE+ on CIFAR-10 and CIFAR-100 throughout the training phase. The employed backbone is ResNet-18. The x-axis denotes the progression in training epochs, while the y-axis represents different metrics in each column. The first column corresponds to the GaussianNLL loss, the second column to the CrossEntropy loss, the third column is the training accuracy, and the final column is the evaluation accuracy.}
    \label{Fig:Losses}
    
\end{figure}

\paragraph{Diffusion Coverage}
The evolution of diffusion coverage during the training process is depicted in~\cref{Fig:Scale}. 
A discernible trend emerges from the data: 
as the model converges, the diffusion coverage contracts and gradually stabilizes at a value exceeding 0. 
This contraction signifies that the model's representation of augmented data is progressively aligning more closely with the representation of the original data, thus indicating an increase in model smoothness across semantically similar areas of entire data space. 

\paragraph{Loss and Accuracy}
The progression of both loss and accuracy throughout the training process is illustrated in~\cref{Fig:Losses}. Both the GaussianNLL loss and CrossEntropy loss display a steady decline, simultaneously the training and evaluation accuracy exhibit a continual ascension. 
This parallel trend affirms the model's progressive learning efficiency and its growing ability to generalize from the training to the evaluation phase. 

\subsection{Analysis for Diffusion Coverage of Unseen Distributions} \label{APP:Prob}

To further validate our generalizability on completely unobserved distribution, we conduct experiments concerning the
diffusion coverage of unseen distribution, i.e., probability of samples under training diffusion distribution, where the sample is augmented by training-unseen augmentations. The experiment is based on the 2-$\sigma$ rule of the Gaussian distribution. The implementation proceeds as follows:  

During the training phase, we used a single augmentation method $\rm{op_{train}}$, while a distinctive technique $\rm{op_{test}}\neq \rm{op_{train}}$ was used during testing. Given a model $f$ trained on $\rm{op_{train}}$ and a sample $x$. We apply $\rm{op_{test}}$ to $x$ to derive an augmented sample $x'$. Inputting $x$ into f gives $\mu_l$ and $\sigma_l$, the representation and the diffusion range of layer $l$. 
Inputting $x'$ into f gives $h_l$, the representation of layer $l$. For each layer, we judge whether $h_l$ is a normal sample under the distribution $N(\mu_l,\sigma_l)$ based on the 2-$\sigma$ rule of the Gaussian distribution. The distance-sigma ratio $\frac{\lvert h_l-\mu_l \lvert }{\sigma_l}$ represents the degree of normal sampling. A smaller value indicates a higher probability that $h_l$ is a normal sample under the distribution $N(\mu_l,\sigma_l)$. If the ratio exceeds 2, it suggests that the sample has less than a 5\% chance of being normal and is thus likely to be an outlier.

We do not directly calculating the probability $p(h_l;\mu_l,\sigma_l)$ due to the continuous nature of our distribution, which encompasses an infinite sample space. For any specific sample, the probability approximates zero. Therefore, given the probability density function $d(x)$ and the specific sample $h_l$, we can only obtain the probability density $d(h_l)$, reflecting the relative density compared to other samples rather than an absolute probability. As such, the most effective method to ascertain whether $h_l$ can be generated by this distribution is based on the 2-$\sigma$ rule. 

The complete changing curve of the ratio $\frac{\lvert h_l-\mu_l \lvert}{\sigma_l}$ throughout the training process is provided in \cref{Fig:Ratio}. Each row represents a distinct type of training augmentation: brightness, Gaussian, and rotation from top to bottom. Each column is corresponding to each layer of neural network. Each sub-figure includes 3 lines corresponding to the changing trends of ratios during training for test samples generated by 3 different test augmentations. The red dashed line marks the position where ratio is 2

As the training gradually stabilizes, two observations can be noted: (1) Even when training is conducted on a single augmentation completely unseen during the testing phase, the test-phase augmentation samples still have a high likelihood of being normal samples within the model distribution, as they fall within 2-$\sigma$ or even smaller. (2) Different augmentation techniques offer varying capabilities of covering unseen distributions. Of the three augmentations experimented in this study, rotation demonstrated the strongest capability, as it swiftly achieved a lower ratio.

\section{Settings}
\label{App:Settings}
This section provides a detailed description of the experimental setups and configurations adopted in this study. These include: 
(1) Datasets summarized in~\cref{App:Datasets}; 
(2) Baseline models against which we benchmarked our proposed model, as is introduced in~\cref{App:Baselines}; 
(3) Metrics employed to gauge performance, as is introduced in~\cref{App:Metrics};
(4) Hyper parameters selected to optimize our model, as is summarized in~\cref{App:Hyper_Parameters}.
(5) Computing Resources summarized in~\cref{App:Resources}.

\subsection{Datasets} 
\label{App:Datasets}
We perform experiments on 7 common image classification datasets including CIFAR-10, CIFAR-100, Tiny-ImageNet, CIFAR-10-C, CIFAR-100-C, Tiny-ImageNet-C and PACS.

\paragraph{Dataset of Original Distribution}
Nature distribution of CIFAR-10, CIFAR-100~\cite{krizhevsky2009learning} and Tiny-ImageNet\cite{le2015tiny} are datasets of original(clean) distribution. CIFAR-10 and CIFAR-100 datasets consist of 60,000 color images, each of size 3x32x32 pixels. CIFAR-10 is categorized into 10 distinct classes with 6000 images per class. CIFAR-100 is more challenging, as these images are distributed across 100 classes, with 600 images per class. 
Tiny-ImageNet datasets consist of 110,000 color images, each of size 3x64x64 pixels, which are categorized into 200 distinct classes with 550 images per class.
Both CIFAR-10 and CIFAR-100 are subdivided into a training set of 50,000 images and a test set of 10,000 images. Tiny-ImageNet
is subdivided into a training set of 100,000 images and a test set of 10,000 images.

\paragraph{Dataset of Corrupted Distributions}
CIFAR-10-C, CIFAR-100-C and Tiny-ImageNet-C\cite{hendrycks2018benchmarking} are variants of the original CIFAR-10, CIFAR-100 and Tiny-ImageNet datasets that have been artificially corrupted into 19 types of corruptions at 5 levels of severity, resulting in 95 corrupted versions of the original test set images. 
The corruptions include 15 main corruptions: Gaussian noise, shot noise, impulse noise, defocus blur, glass blur, motion blur, zoom blur, snow, frost, fog, brightness, contrast, elastic, pixelation, and JPEG. Both datasets also contain 4  corruptions that are not commonly used: speckle noise, Gaussian blur, spatter, saturation. 
All these corruptions are simulations of shifted distributions that models might encounter in real-world situations.

\paragraph{Datsset of PACS}
PACS\cite{li2017deeper} is an image dataset popular used in domain generalization and transfer learning. It consists of 4 domains, namely Photo (1,670 images), Art Painting (2,048 images), Cartoon (2,344 images) and Sketch (3,929 images). Each domain contains 7 categories.

\begin{table}[ht]
\centering
\caption{Summary of Original \& Corruption Datasets}
\setlength{\tabcolsep}{2mm}
\begin{tabular}{lccccc}
\toprule
\textbf{Dataset} & \textbf{\#Train} & \textbf{\#Test} & \textbf{\#Corr.} & \textbf{\#Severity} & \textbf{\#Class.} \\
\midrule
CIFAR-10 & 50,000 & 10,000 & 1 & 1 & 10  \\
CIFAR-100 & 50,000 & 10,000 & 1 & 1 & 100  \\
Tiny-ImageNet & 100,000 & 10,000 & 1 &1 & 200\\
CIFAR-10-C & - & 950,000 & 15(+4) &5 & 10 \\
CIFAR-100-C & - & 950,000 & 15(+4) &5 & 100  \\
Tiny-ImageNet-C &-& 750,000 & 15&5 & 200\\
\bottomrule
\end{tabular}
\label{Tab:Dataset_Corr}
\end{table}

\begin{table}[h!]
\centering
\caption{Summary of PACS Datasets}
\setlength{\tabcolsep}{6mm}
\begin{tabular}{lccccc}
\toprule
\textbf{Domain} & \textbf{\#Sample} & \textbf{\#Class} & \textbf{Size} \\
\midrule
\textbf{P}hoto & 1,670 & 7 &3x227x227 \\
\textbf{A}rt & 2,048 & 7 & 3x227x227\\
\textbf{C}artoon & 2,344 & 7 & 3x227x227\\
\textbf{S}ketch & 3,929 & 7 & 3x227x227\\
\bottomrule
\end{tabular}
\label{Tab:Dataset_PACS}
\end{table}

\subsection{Baselines}
\label{App:Baselines}

The various baseline models we selected for comparison with our PDE+ can be categorized into five distinct groups, each representative of a specific approach or methodology:

\paragraph{Standard Training} This category is represented by the foundational ResNet-18 architecture~\cite{he2016deep}. It serves as a reference point for the rest of the methodologies, showcasing the results of a model trained without any specific regularization, noise injection, or augmentation.

\paragraph{Lipschitz Continuity} Methods within this category impose model smoothness directly via the Lipschitz continuity principle. For instance, the work presented in ~\cite{165600} achieves this smoothness by implementing gradient regularization. This approach enhances the stability of the model against small changes in the input space.

\paragraph{Noise Injection} This category encompasses methods that utilize various strategies to inject noise. EnResNet~\cite{doi:10.1137/19M1265302} improves performance through an ensemble of ResNets with injected noise. RSE~\cite{Liu_2018_ECCV} presents a framework that injects noise and employs self-ensembling during testing to enhance model robustness. NFM~\cite{lim2022noisy} combines noise injection and Manifold Mixup~\cite{verma2019manifold} to expand the generalization capacity of the model.

\paragraph{Data Augmentation} This group consists of methods via various augmentation strategies to diversify and expand the training dataset.
Gaussian noise creates minor variations to increase the robustness.
Mixup~\cite{zhang2018mixup} generates synthetic training examples by creating linear interpolations of random pairs of images and their corresponding labels 
DeepAug~\cite{9710159} uses a deep generative model to create synthetic training examples, increasing the diversity of the training set.
AutoAug~\cite{Cubuk_2019_CVPR} utilizes reinforcement learning to discover the best augmentation policies from a search space of possible augmentations, optimizing the model's validation accuracy.
AugMix~\cite{hendrycks*2020augmix} employs a combination of multiple augmentation transformations and creates a mixture of augmented images, helping the model to generalize better by exposing it to more varied and complex examples.

\paragraph{Adversarial Training} This category uses adversarial examples during training to bolster generalization and robustness.
PGD~\cite{madry2018towards} generates adversarial examples using an iterative process, thereby strengthening the model's robustness against adversarial attacks, which has been recently discovered that by judicious selecting the perturbation radius, it can enhance non-adversarial generalization on common corruptions~\cite{kireev2022on}.
RLAT~\cite{kireev2022on} introduces an efficient relaxation to AT via the distance metric of learned perceptual image patch similarity. This approach combined with data augmentation methods can achieve state-of-the-art performance on common corruptions.

All methods employed for comparison are representative within their respective categories and represent state-of-the-art baselines, providing a comprehensive range of approaches against which the performance of our PDE+ model can be evaluated.

\subsection{Metrics}
\label{App:Metrics}

Three metrics are typically employed for reporting the performance of a model on CIFAR-10-C and CIFAR-100-C across  different corruption types and severity levels: Average Accuracy, Mean Corruption Error (mCE)~\cite{hendrycks2018benchmarking}, and Relative Mean Corruption Error (rmCE)~\cite{hendrycks2018benchmarking}. These metrics provide a comprehensive evaluation of a model's generalization in handling diverse corruptions and severities, thereby offering a multi-faceted perspective on model performance.

\paragraph{Average Accuracy} Average accuracy is the accuracy averaged over all severity levels and corruptions. Consider there are a total of $C$ corruptions, each with $S$ severities. For a model $f$, let $\mathcal{E}_{s,c}(f)$ denote the top-1 error rate on the corruption $c$ with severity level $s$ averaged over the whole test set,
\begin{equation}
    \mathrm{Accuracy}_f=1-\frac{1}{C \cdot S} \sum_{c=1}^{C} \sum_{s=1}^S \mathcal{E}_{s, c}(f).
\end{equation}

\paragraph{Mean Corruption Error} Mean corruption error (mCE) is a metric used to measure the performance improvement of model $f$ compared to a baseline model $f_0$. We use ResNet-18 as the baseline model, instead of AlexNet~\cite{krizhevsky2017imagenet} which is traditionally used in~\cite{hendrycks2018benchmarking}.
We define $\mathrm{mCE}_f$ as follows,
\begin{equation}
\mathrm{mCE}_f=\frac{1}{C} \sum_{c=1}^C \frac{\sum_{s=1}^S \mathcal{E}_{c, s}(f)}{\sum_{s=1}^S \mathcal{E}_{c, s}\left(f_0\right)} .
\end{equation}

\paragraph{Relative Mean Corruption Error}
Relative mean corruption error (rmCE) is a variant to mCE, which  takes the error rate of models trained on natural data distribution into consideration.
Denote the error rate of models $f$ trained on natural data distribution into consideration as $\mathcal{E}_{\mathrm{nat}}(f)$ and the error rate of models $f_0$ trained on natural data distribution into consideration as $\mathcal{E}_{\mathrm{nat}}(f_0)$.
We define $\mathrm{rmCE}_f$ as follows,
\begin{equation}
\mathrm{rmCE}_f=\frac{1}{C} \sum_{c=1}^{C} \frac{\sum_{s=1}^S \mathcal{E}_{s, c}(f)-\mathcal{E}_{\mathrm{nat}}(f)}{\sum_{s=1}^S \mathcal{E}_{s, c}(f_0)-\mathcal{E}_{\mathrm{nat}}(f_0)} .
\end{equation}

\subsection{Hyper-parameters}
\label{App:Hyper_Parameters}

In this section, we outline the hyperparameters chosen for our experiments, which are based on empirical evaluations. These settings enable the reproducibility of the results presented in our study.

\begin{table}[!ht]
\renewcommand{\arraystretch}{0.5}
\caption{Summary of Hyper-parameters}
\centering
\begin{small}
\setlength{\tabcolsep}{2mm}
\begin{tabular}{lccccccc}
\toprule
\multirow{2}*{Data} & \multirow{2}*{Epochs} & \multirow{2}*{B.S.}& \multicolumn{3}{c}{Classifier} &\multicolumn{2}{c}{Diffuser} \\
\cmidrule(lr){4-6} \cmidrule(lr){7-8}
& & & lr & Opt. & Sch. & lr & Opt. \\
\midrule
CIFAR-10 & 200 & 128 & 0.05 & SGD & CosineLR & 0.015 & Adam  \\
CIFAR-100 & 200 & 128 & 0.05 & SGD & CosineLR & 0.010 & Adam  \\
Tiny-ImageNet & 200 & 128 & 0.06 & SGD & CosineLR & 0.005 & Adam  \\
PACS & 200 & 128 & 0.02 & SGD & CosineLR & 0.005 & Adam  \\
\bottomrule
\end{tabular}
\end{small}
\end{table}

\subsection{Computing Resources}
\label{App:Resources}
All our experiments are performed on RedHat server (4.8.5-39) with Intel(R) Xeon(R) Gold 5218 CPU $@$ 2.30GHz4 and $4 \times$ NVIDIA Tesla V100 SXM2 (32GB)

\section{Limitations and Future Explorations } \label{App:Limitation}

\paragraph{Limitations and Future Works}
This paper presents two limitations that need to be addressed in future work. Firstly, the implementation is based on residual connected network due to the form of TE, which can be further explored to structures without residual connection. Secondly, we acknowledge that the Gaussian prior used for the implementation can be replaced with other priors, such as Laplace prior~\cite{kotz2001laplace}, which is worthy of further exploration. Lastly, experimental on adversarial samples is worth exploring.

\paragraph{Broader Impacts}
Our method does not raise concerns regarding negative societal impact, as it primarily focuses on enhancing the generalization performance of neural networks. By improving generalization in unknown real-world scenarios, this approach can contribute to the development of more reliable and trustworthy machine learning models.

\end{document}